\pdfoutput=1

\documentclass[11pt]{article}

\usepackage[final]{acl}

\usepackage{algorithm}
\usepackage{algpseudocode}

\usepackage{times}
\usepackage{latexsym}
\usepackage{booktabs}
\usepackage{multirow}
\usepackage{graphicx}
\usepackage{xspace}
\usepackage{enumitem}
\usepackage{xcolor}
\usepackage{pifont}
\usepackage{amsmath}
\usepackage{subfigure}

\usepackage[T1]{fontenc}

\usepackage[utf8]{inputenc}

\usepackage{microtype}

\usepackage{inconsolata}

\usepackage{graphicx}
\usepackage{microtype}
\usepackage{textcomp}
\usepackage{multicol}
\usepackage{amssymb}
\usepackage{adjustbox}
\usepackage{wrapfig}

\newcommand{\cmark}{\ding{51}}%
\newcommand{\xmark}{\ding{55}}%

\def \name{\textsc{PfP}\xspace}

\title{Debiasing Online Preference Learning via Preference Feature Preservation}

\author{
  Dongyoung Kim$^{1}$, Jinsung Yoon$^{2}$ , Jinwoo Shin$^{1}$, Jaehyung Kim$^{3}$\\
  $^{1}$KAIST, $^{2}$Google Cloud AI Research, $^{3}$Yonsei University  \\ \texttt{kingdy2002@kaist.ac.kr},~~  \texttt{jaehyungk@yonsei.ac.kr}
  }

\begin{document}
\newcommand{\fix}{\marginpar{FIX}}
\newcommand{\new}{\marginpar{NEW}}

\maketitle

\begin{abstract}
Recent preference learning frameworks for large language models (LLMs) simplify human preferences with binary pairwise comparisons and scalar rewards.
This simplification could make LLMs' responses biased to mostly preferred features, and would be exacerbated during the iterations of online preference learning steps. 
To address these challenges, we propose a novel framework coined \name{} (\textbf{P}reference \textbf{F}eature \textbf{P}reservation). 
The key idea of \name{} is maintaining the distribution of human preference features and utilizing such rich signals throughout the online preference learning process.
Specifically, \name{} first extract preference features from offline pairwise human preference data and trains a feature classifier.
Then, using trained classifier and the distribution preserving optimization, \name{} maps appropriate preference features for a new input instruction during online learning. 
Lastly, \name{} trains LLM using the existing preference learning method, by incorporating the preference feature into system prompts and enabling LLM to explicitly handle various human preferences.  
Our experiments demonstrate that \name{} successfully mitigates the bias in preference features during online learning, and hence achieves superior performance compared to previous preference learning methods on standard benchmarks to evaluate LLM alignment.\footnote{\url{https://github.com/kingdy2002/PFP}} 

\end{abstract}

\section{Introduction}
\begin{figure*}[t]
\vspace{-0.2in}
\begin{center}
    {
    \subfigure[Biased feature distribution]
        {
        \includegraphics[width=0.31\textwidth]{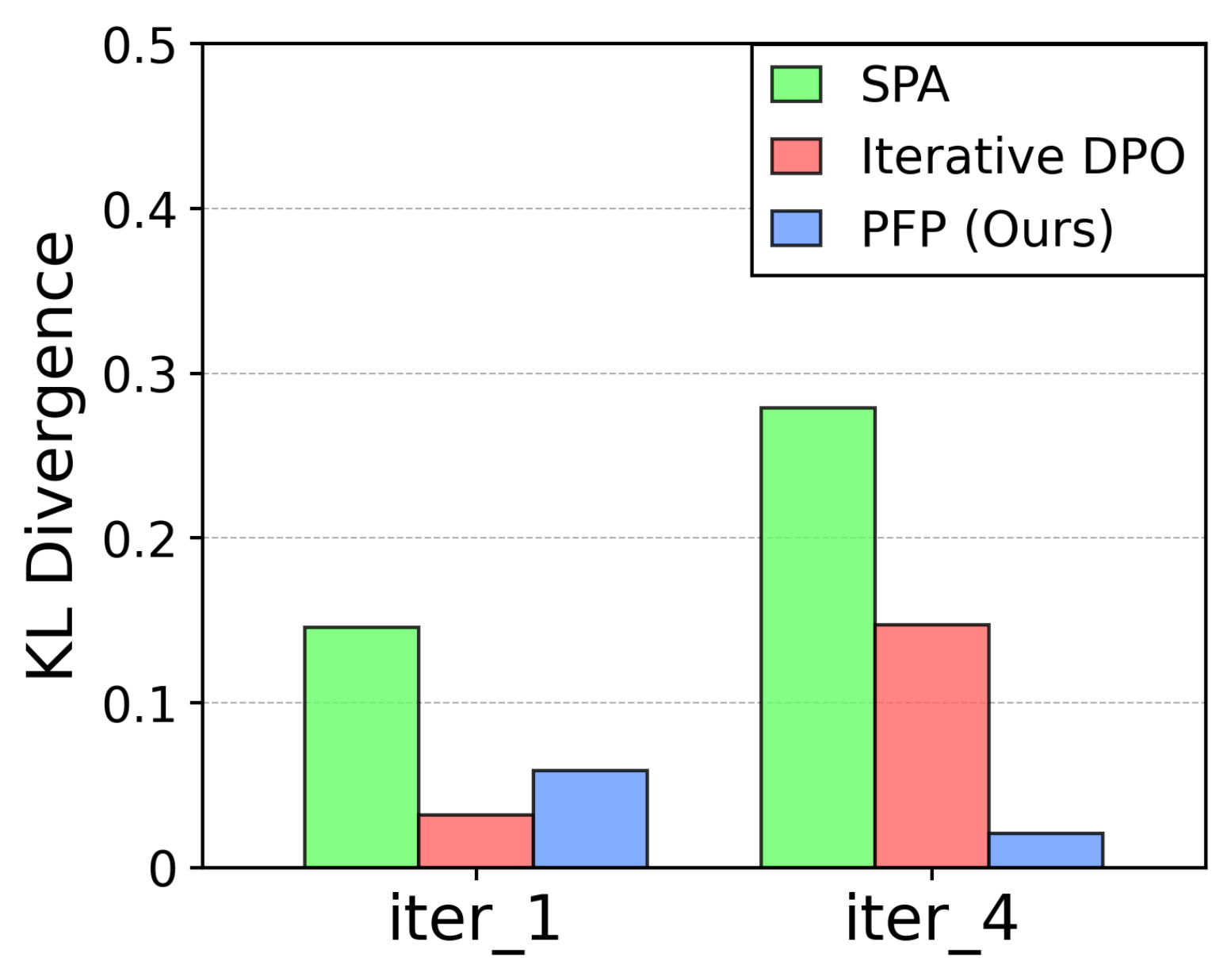}
        \label{fig:bias}
        } 
    \subfigure[System prompt from features]
        {
        \includegraphics[width=0.31\textwidth]{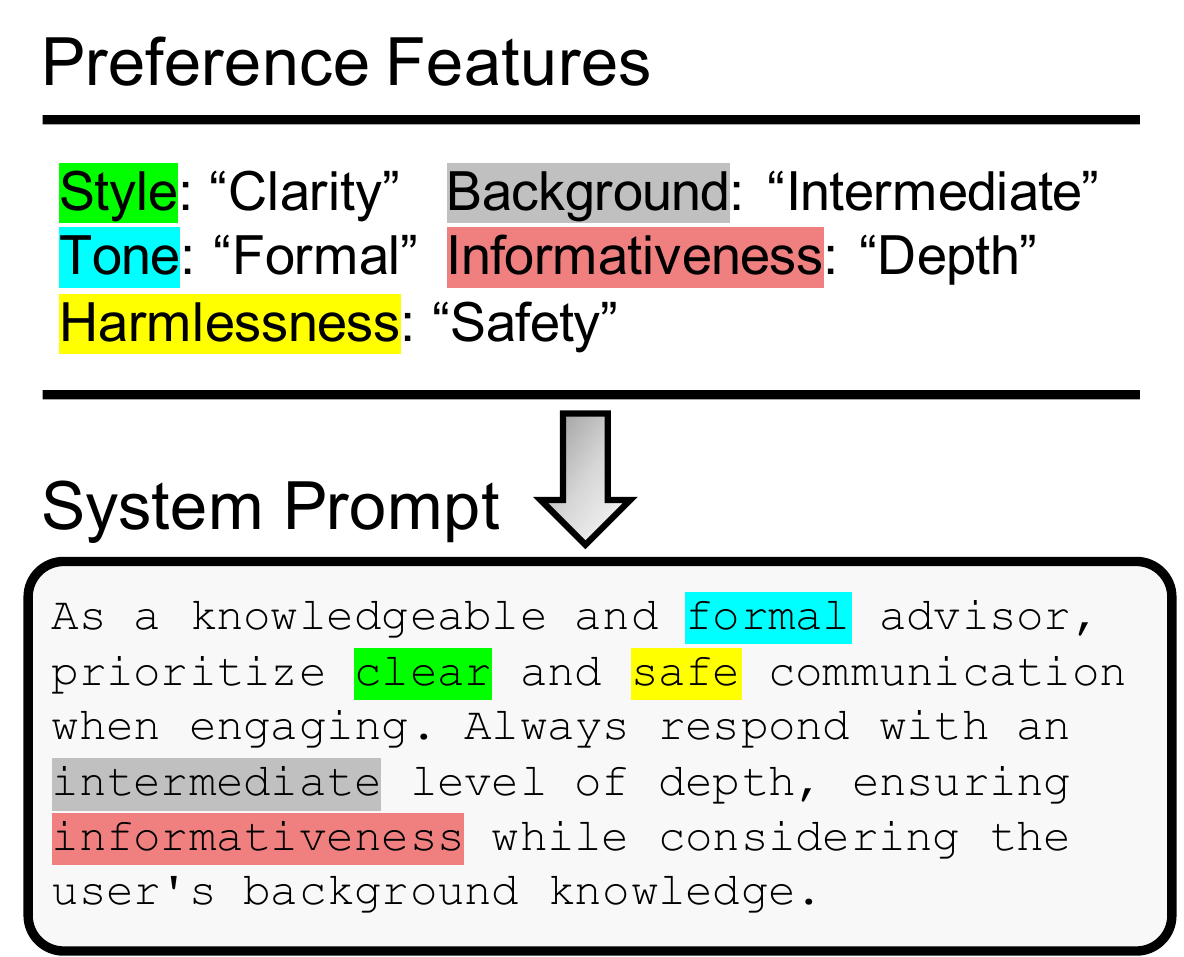}
        \label{fig:example}
        }
    \subfigure[Effectiveness of debiasing via \name{}]
        {
        \includegraphics[width=0.31\textwidth]{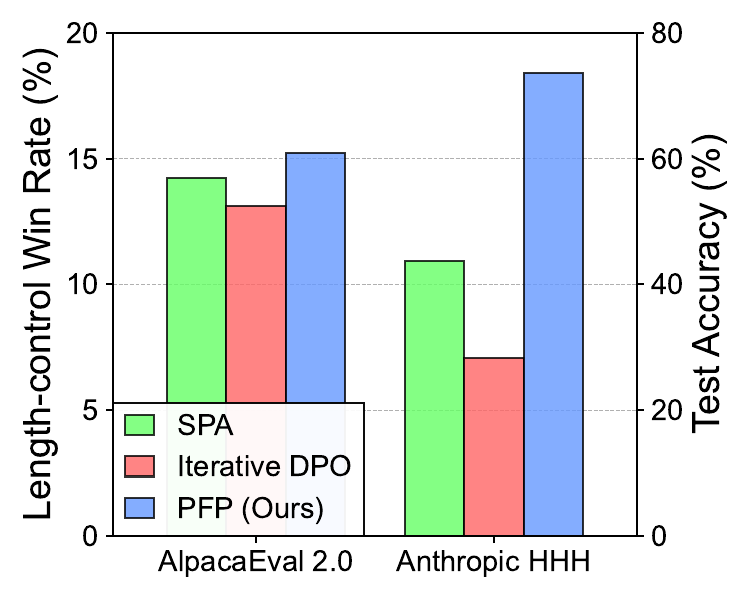}
        \label{fig:res}
        }
    }
\end{center}
\vspace{-0.15in}
\caption{{\textbf{Motivation for debiasing online preference learning.}
(a) For existing online preference learning methods (SPA, Iterative DPO), the preference feature distribution of trained models, obtained by inversely asking GPT-4o, progressively diverges from the original preference distribution, captured by the initial DPO model.
(b) To preserve the distribution, we propose to map each input instruction with the specific preference features and then convert it into the system prompt to enable LLM to explicitly handle preference features and preserve them. 
(c) We demonstrate that our approach, \name{}, achieves strong performance not only on general preference benchmarks such as AlpacaEval2.0 but also on Anthropic-HHH, which incorporates critical metrics like harmlessness and honestness. 
}}
\label{fig:motivation}
\vspace{-0.1in}
\end{figure*}

Aligning large language models (LLMs) using human feedback, particularly by learning from human preferences, yields remarkable successes in various NLP tasks and real-world applications such as coding assistants and chatbots \citep{claude3.5,dubey2024llama,openai2024gpto1,team2023gemini}.
To improve the alignment of LLMs, various preference learning algorithms, such as Reinforcement Learning from Human Feedback (RLHF) \citep{ouyang2022training} and Direct Preference Optimization (DPO) \citep{rafailov2023direct}, have been explored.
A common assumption across these works is that human preference is provided in a binary pair-wise comparison \citep{ziegler2019fine, hong2024orpo}. 
This approach enables easy modeling of human preference using the scalar reward such as the Bradley-Terry (BT) model \citep{bradley1952rank}.
However, human preference is determined by several underlying features \citep{li2024dissecting, oh2024uncovering}, and hence such simplification has critical limitations and fails to capture the complexity of human preferences. 
For example, even though human preference can be varied depending on the preference feature, the most dominant preference feature would be only considered to determine the binary human preference label.  
This issue becomes even more problematic in \textit{online preference learning} scenarios, which progressively improves the alignment of LLMs by iterating the generation of preference data and learning from them \citep{xiong2024iterative, wu2024self, rosset2024direct}.
During online preference learning, LLM will generate responses biased toward specific preference features, and the preference annotators, such as the external reward model \citep{jiang2023llm}, will provide positive feedback on this. 
As such iterations go on, the bias of LLM accumulates (see Fig.~\ref{fig:bias}), and hence it results in the reduced diversity and quality of LLM's responses.

\noindent\textbf{Contribution.}  
To address these challenges, we propose a novel online preference learning framework called \name{} (\textbf{P}reference \textbf{F}eature \textbf{P}reservation). 
Our approach is to ensure that the distribution of preference features remains consistent throughout the online preference learning process. 
Here, the key idea is to explicitly extract preference features of each input instruction and handle them using system prompts of LLMs (see Fig.~\ref{fig:example}); it enables LLMs to generate and learn preference data with intent. 
Specifically, \name{} first estimates the initial distribution of preference features of the given human preference dataset, by inferring which features mainly determine binary human preferences. 
We then train a preference feature classifier, which maps each input instruction to appropriate preference features with additional optimization for the distribution preservation, during the online learning process. 
Finally, \name{} trains LLM using the existing preference learning framework, by converting the mapped preference features of each generated data into the system prompts of LLMs.

We demonstrate the effectiveness of the proposed \name{} by applying it to align recent open-sourced LLMs with the commonly used preference dataset, UltraFeedback \cite{cui2023ultrafeedback}.
The experimental results demonstrate that \name{} successfully prevents LLMs being biased to specific preference features during online learning, and results in the improved alignment in various aspects (Fig.~\ref{fig:res}). 
For example, our framework achieves 7.66\% increase in AlpacaEval 2.0 length-controlled win rate compared to the SFT model. 
Also, \name{} achives 2.11\% larger increase compared to Iterative DPO, an online preference learning method with external reward model.
In addition, unlike other baselines for online preference learning, \name{} successfully improves harmlessness and honestness of the responses \cite{askell2021general} through multiple iterations.  
More interestingly, \name{} exibits additional advantage that reduces the occurrence of length bias during online preference learning, despite not being specifically designed to address this. 
Overall, these results demonstrate that \name{} is highly effective and practical for real-world applications, and underscoring the importance of debiasing to learn human preference for LLM alignment. 
\section{Related Works}
\textbf{LLM alignment with human preference.} 
\citep{ziegler2019fine, ouyang2022training}.
Aligning LLMs with human intentions and values using human feedback data now becomes a defacto standard to obtain well-performing LLMs \citep{ziegler2019fine, ouyang2022training}.
Typically, this feedback is collected by asking human annotators to compare two responses generated from the same input prompt and assign a binary preference label.
One of the most widely adopted approaches is RLHF \citep{christiano2017deep, stiennon2020learning}, where a reward model is trained to model human preferences \citep{bradley1952rank}, and LLM is then fine-tuned to optimize for this learned reward. 
To prevent issues such as reward over-optimization and model collapse, KL divergence regularization is commonly employed during this process.
However, RLHF presents several challenges such as computational overheads and the training instability.
To address these issues, alternative approaches have been extensively proposed \citep{rafailov2023direct, zhao2023slic, meng2024simpo, hong2024orpo}; for instance, DPO \citep{rafailov2023direct} eliminates the need for a separate reward model by deriving a training objective that is mathematically equivalent to RLHF.

\noindent\textbf{Online preference learning.} 
Existing preference learning methods can generally be categorized into two approaches depending on whether they use the fixed human preference dataset (\textit{offline preference learning}, \textit{e.g.}, DPO) or progressively enlarge dataset from the iterations of sampling and labeling (\textit{online preference learning}, \textit{e.g.}, RLHF). 
While online methods typically achieve superior performance due to train with more data, they also demand more computations from sampling responses and labeling preferences.
To address this challenge, recent work has focused on developing efficient batch-online preference learning methods, such as Iterative DPO \citep{xu2023some, xiong2024iterative, rosset2024direct, wu2024self,  calandriellohuman2024}. 
Iterative DPO generates thousands of responses in each iteration (batch) and constructs labeled preference datasets by judging the preference using the reward model \citep{jiang2023llm}. 
This dataset is then used to train LLMs with offline methods like DPO, and the iteration repeats, resulting in more efficient and stable alignment. 
\section{Preliminary}\label{sec:3}
Let the LLM policy be denoted as $\pi_{\theta}$, which can generate output sequence (\textit{i.e.} response) $y$, given input sequences composed of \textit{system prompt} $s$ and \textit{instruction} $x$, \textit{i.e.}, $y \sim \pi_{\theta}(s,x)$.
Here, the system prompt $s$ is usually considered to be fixed regardless of the input instruction $x$.
For convenience, we assume that $s$ is always included as the input of $\pi_{\theta}$ and hence omit $s$ in the equations in the below parts. 
Next, we assume that we have the labeled preference dataset, $\mathcal{D}=\{(x,y_l,y_w)\} $, where $y_l$ and $y_w$ are the dis-preferred and preferred responses for the corresponding instruction $x$, respectively. 

\noindent\textbf{RLHF and DPO.} 
To train $\pi_{\theta}$ with $\mathcal{D}$ for the alignment, RLHF first introduces the reward model $r(x,y)$ which can convert human preference data into scalar values. 
Specifically, the reward model $r(x,y)$ is often modeled with the Bradley-Terry (BT) model \citep{bradley1952rank}, and then it can yield the probability $p(y_{w} \succ y_{l} \mid x)$ that response $y_{w}$ is preferred over $y_{l}$ as follow: 
\begin{equation*}
p(y_{w} \succ y_{l} \mid x) = \frac{\exp\left(r(x, y_w)\right)}{\exp\left(r(x, y_w)\right) + \exp\left(r(x, y_l)\right)}.
\end{equation*}
As the optimal reward function $r(x,y)$ is not accessible, a parameterized reward model $r_{\phi}(x,y)$ is usually introduced by optimizing its parameters with the maximum-likelihood objective on the preference dataset. 
With this reward model, RLHF optimizes LLM $\pi$ to maximize this reward with the additional regularization of the KL divergence between the current policy and the reference policies ($\pi_\text{ref}$) to prevent reward over-optimization:
\begin{align*}
    \mathcal{L}_\text{\tt RLHF}=
    & -\mathbb{E}_{y \sim \pi_{\theta}, x \sim \rho} \left[ r_{\phi}(x,y) \right] \\ &
    + \beta \mathrm{D}_{\mathrm{KL}} \left( \pi_{\theta}(y|x) \parallel \pi_{\text{\tt ref}}(y|x) \right).
\end{align*}

To remove the necessity of the reward model in RLHF, DPO proposed a method that is mathematically equivalent to the original RLHF objective and can directly optimize the internal reward modeled by LLM $\pi$ itself, by maximizing the weighted likelihood gap between $y_w$ and $y_l$:
\begin{align}\label{eq:selfee_pref}
    p_{\theta}(y_{w} \succ y_{l} | x) =&  \sigma (\beta \log \frac{\pi_{\theta}(y_w|x)}{\pi_{\text{\tt ref}}(y_w|x)} \nonumber\\ & - \beta \log \frac{\pi_{\theta}(y_l|x)}{\pi_{\text{\tt ref}}(y_l|x)}).
\end{align}
\begin{equation*}
    \mathcal{L}_\text{\tt DPO} = \mathbb{E}_{(x, y_w, y_l) \sim \mathcal{D}} \left[-\log p_{\theta}(y_{w} \succ y_{l} | x) \right].
\end{equation*}

\noindent\textbf{Online preference learning and SPA.} 
In the online preference learning scenario, we have unlabeled instruction datasets $X_{t}=\{x\}, ~ t=1,...,T$ where $X_{t} \cap X_{t'} = \emptyset$ when $t \ne t'$.
For $t$-th iteration, the preference dataset { $\mathcal{D}_{t}=\{(x,y_l,y_w)|x \in X_{t}\}$} is constructed by (1) sampling two responses for each instruction $x \in X_{t}$ using LLM policy $\pi_{t-1}$ from the previous iteration, \textit{i.e.}, $y_1,y_2 \sim \pi_{t-1}(x)$, and (2) judging the preference between them. 
Then, LLM policy $\pi_{t}$ which is initialized with $\pi_{t-1}$ is trained with $\mathcal{D}_{t}$ using the existing preference learning method. 
One representative approach is Iterative DPO \citep{xu2023some}, where the external reward model is used for the preference judgments and $\pi_{t}$ is trained with $\mathcal{D}_{t}$ using DPO. 
Since choosing the proper reward model is non-trivial in our problem, we adopt SPA \citep{kim2025debiasing} as the online preference learning algorithm. 
SPA conducts preference labeling using the implicit reward derived from the DPO's objective function (Eq.~\ref{eq:selfee_pref}), unlike the other online preference learning methods using the external reward model:
\begin{align}\label{eq:selfee_label}
    & (y_{w}, y_{l}) = (y_{1}, y_{2})~\text{ if }~p_{\theta_{t-1}}(y_{1} \succ y_{2} | x) > 0.5 \nonumber \\ & (y_{w}, y_{l}) = (y_{2}, y_{1})~~\text{else}~
\end{align}
\section{\name{}: Debiased LLM Alignment via Preference Feature Preservation}\label{sec:4}

\begin{figure*}[t]
    \centering
    \vspace{-0.2in}
    \includegraphics[width=0.85\textwidth]{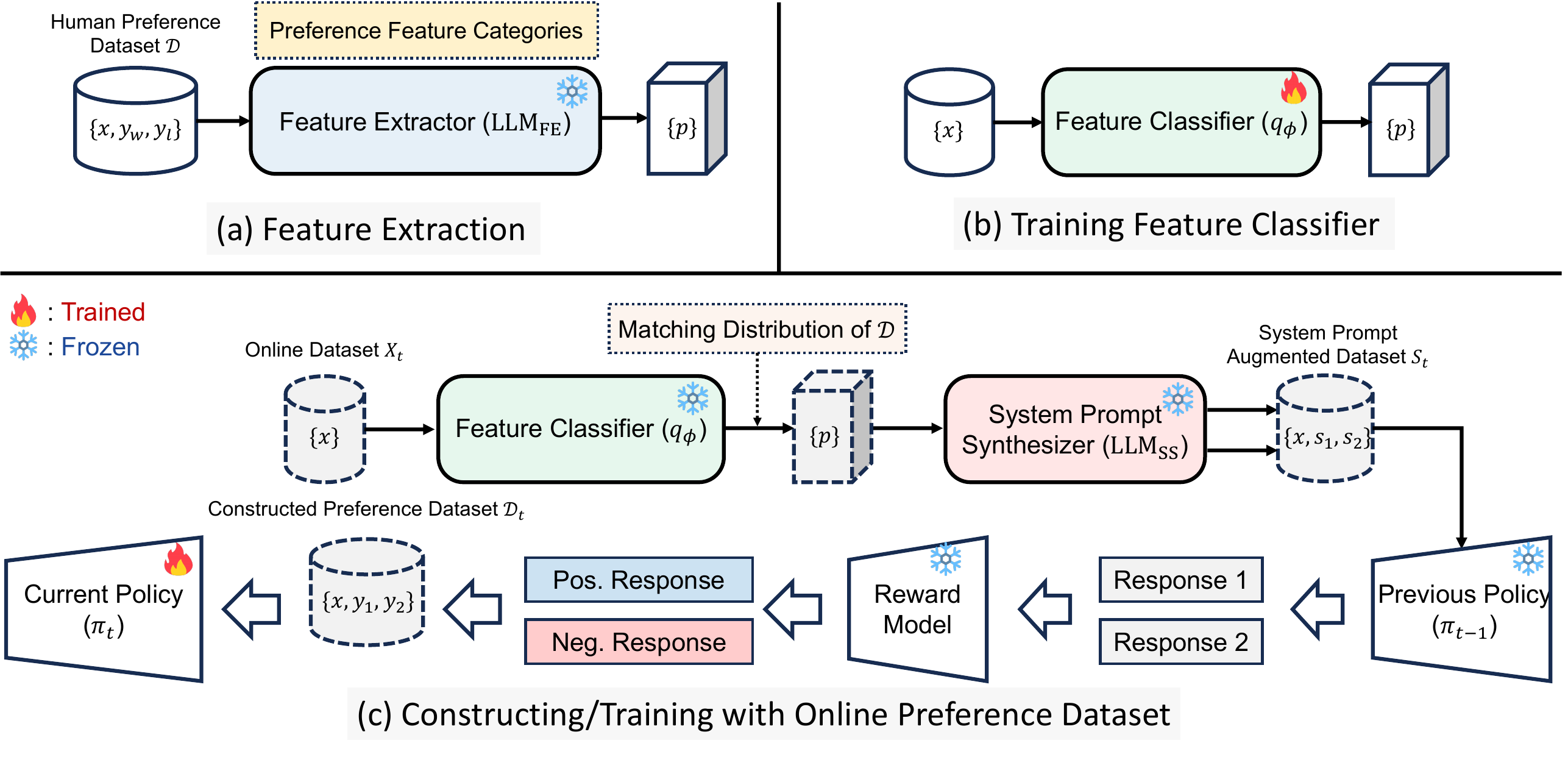} 
    \vspace{-0.05in}
    \caption{\textbf{Illustration of the proposed \name{} framework.} (a) \name{} first extracts the preference feature among the predefined categories for a given human preference dataset using an LLM-based feature extractor. (b) With the extracted features, \name{} trains the feature classifier. (c) The trained feature classifier along with am additional distribution adjustment step assigns the preference feature for a new instruction in a online data. Then, the LLM-based system prompt synthesizer converts it into two system prompts, where each system prompt is used to sample the separate response. Then, the labeled preference dataset is constructed and the current policy LLM is trained.} 
    \label{fig:illustration}
    \vspace{-0.1in}
\end{figure*}

In this section, we present \name{} (  \textbf{P}reference \textbf{F}eature \textbf{P}reservation) to align LLM by reducing the bias during online preference learning. 
Our main idea is to explicitly extract preference features of input instruction, and handle them using system prompts of LLM. 
The overview of \name{} is presented in Fig.~\ref{fig:illustration}.

\subsection{Extracting preference feature from binary human preference data}\label{sec:4.1}
We first assume that some underlying keys features mainly determine the human preference between responses for a given input prompt; we call them as \textit{preference features}.
Following \citet{lee2024aligning}, we define a five-dimensional preference feature where each dimension captures different domain of preference: \textit{style, tone, harmlessness, background knowledge, and informativeness}.
Then, each dimension contains five sub-features; for example, \textit{style} dimension consists of following five sub-features: \textit{clarity, conciseness, format, vividness, and consistency} (see full description in Table~\ref{tab:featureset}).
Under this definition, we extract the preference features of the pairwise offline human preference data $\mathcal{D}$ using the feature extractor.
We implement the feature extractor by prompting LLM such as GPT-4o \citep{openai2024gpto1}, to infer the likely preference features that led the annotators to provide a given preference label. 
Specifically, for the input instruction $x$ and the two responses $y_{w}$ and $y_{l}$, the feature extraction yields $\mathbf{p} = \text{LLM}_{\text{FE}}(x,y_l,y_w)$ where $\mathbf{p}=[p_{1},...,p_{5}]$, where each $p_i$ represents a one-hot label over the 5 sub-features for $i$-th dimension (\textit{i.e.}, $p_i \in [0, 1]^5$ and $\sum_{j=1}^5 p_i^j = 1$). 
Then, the extracted preference features are added to the human preference data $\mathcal{D}$ and it yields  $\mathcal{D}_{\text{FE}}=\{(\mathbf{p},x,y_l,y_w)\}$.
 
\subsection{Distribution preserved feature mapping}\label{sec:4.2} 
To preserve the feature distribution over each iteration of online preference learning, we first map each instruction $x \in X_{t}$ used in online learning to the proper preference features. 
Then, the preference feature distribution can be preserved by explicitly incorporating the assigned features during response generation and preference judgment.  
Specifically, this process involves two key components:  (a) learning feature classifier, and (b) adjusing assigned feature using relabeling technique.

\noindent\textbf{Learning feature classifier.} 
\name{} introduces an auxiliary classifier $q_{\phi}$ to predict appropriate preference features for the given input instruction $x$. 
Specifically, $q_{\phi}$ is trained with a conventional supervised learning with cross-entropy loss, using the input instructions $x$ and the extracted features $p$ in $\mathcal{D}_{\text{FE}}$ (\textit{i.e.}, sequence classification).
After the training, $q_{\phi}$ can provide a probability distribution over preference features for a new instruction $x \in X_{t}$ that will be used in online learning.
For each $i$-th dimension, a separate classifier $q_{\phi^{i}}$ is introduced where $q_{\phi^{i}}(x)=[0,1]^5$ and $\sum q_{\phi^{i}}(x)=1$.

\noindent\textbf{Adjusted output prediction.} \label{sec:Adjusted_relableing}
However, due to the difficulty of the given task from limited data and long-tailed nature, the classifier's prediction can be inaccurate and hence limited to preserve the feature distribution. 
To complement the classifier's predictions be aligned with the distribution of human preferences, \name{} adjusts the predicted probabilities by solving the optimization for this. 
Formally, for each $i$-th preference dimension, the human preference feature distribution is empirically derived from $\mathcal{D}_{\text{FE}}$, \textit{i.e.}, ${P}_i = \sum\nolimits_{\mathbf{p} \in \mathcal{D}_{\text{FE}}} p_i / |\mathcal{D}_{\text{FE}}|$.
Next, the output probabilities for all input instructions in $X_{t}$ under $q_{\phi^{i}}$ is collected to measure the distribution, \textit{i.e.}, ${Q}_i = \sum\nolimits_{x \in {X}_{t}} q_{\phi^{i}}(x) / |{X}_{t}|$.
Then, our goal is to find the adjusted output probability $\widetilde{q}_{i}(x) \in [0,1]^{5},~\sum \widetilde{q}_{i}(x)=1$ for $x \in {X}_{t}$ that yields the identical empirical distribution with ${P}_k$ while minimizing the deviation from the original probability $q_{\phi^{i}}(x)$.
This problem can be formulated as below optimization problem:
\begin{align}\label{eq:opt}
& \min_{q} \text{CE}(q_{\phi^{i}}, q) \quad \text{s.t.} \quad \forall x \in {X}_{t}: q(x) \in [0, 1]^{5}, \nonumber \\& ~\sum_{j=1}^{5} q(x)_j =1, ~ \text{and} ~ \sum_{x \in {X}_{t}} q(x) / |{X}_{t}| = {P}_{i}.
\end{align}
where $\text{CE}$ is a cross-entropy. 
Following the previous works \citep{asano2020self, kim2020distribution}, we solve this problem via efficient Sinkhorn-Knopp algorithm \citep{cuturi2013sinkhorn}. 
With $\widetilde{q}_{k}(i)$ from solving Eq. \ref{eq:opt} with $q_{\phi^{i}}$, we sample the preference feature and augment the online dataset $X_{t}$, \textit{i.e.}, $\widetilde{p}_{i} \sim \widetilde{q}_{i}(x)$ and $\widetilde{X}_{t}=\{(\widetilde{\mathbf{p}},x)|x \in X_{t}, \widetilde{\mathbf{p}}=[\widetilde{p}_1,...,\widetilde{p}_5]\}$.

\subsection{Learn to handle preference features through system prompt}\label{sec:4.3}
\textbf{Synthesizing system prompt.} 
We need to generate responses and judge the preference using the LLM policy $\pi_{\theta}$ conditioned on the given preference feature.
However, it can be difficult as the preference features have the form of short words that are not suitable for LLM. 
To address this, we convert these discretized preference features into the system prompt, which is a natural language description about the preference feature. 
Then, we add it in front of the instruct $x$ as the conventional system prompt. 
Specifically, the system prompt $s$ is created through the system prompt synthesizer, which is implemented by prompting LLM that receives features as input and generates a system prompt, \textit{i.e.}, $s \sim \text{LLM}_{\text{SS}}(\widetilde{p})$.
Then, we incorporate the generated system prompt into the online learning dataset, \textit{i.e.}, ${S}_{t}=\{(s,x)|(\widetilde{\mathbf{p}},x) \in \widetilde{X}_{t}\}$. 
We created the prompt for $\text{LLM}_{\text{SS}}$ by modifying the prompt used in \citet{lee2024aligning} (see Appendix \ref{Prompt_Set}). 
Using ${S}_{t}$, one can perform the existing online preference learning method, such as Iterative DPO.

\noindent\textbf{System prompt sampling and scheduling.} 
While incorporating preference features into LLM using the system prompt enables LLM to understand and handle them better, we observe that conditioning specific system prompts could reduce the diversity between sampled responses.
This reduced diversity makes preference judgment between them difficult and consequently leads to decreased performance (see Table~\ref{tab:ablation_scheduling}). 
To prevent this, we propose to augment the online learning data set $X_{t}$ by sampling two system prompts, \textit{i.e.}, $S_{t}=\{(s_{1},s_{2},x)|x \in X_{t}\}$ and $s_{1},s_{2} \sim \text{LLM}_{\text{\tt SS}}(\widetilde{p})$.
Then, during data construction process, each system prompt is used to sample the different response, \textit{i.e.}, $y_i \sim \pi_{t-1}(s_{i},x)$ where $i=1,2$.
Finally, using Eq. \ref{eq:selfee_pref} and \ref{eq:selfee_label}, we judge the preference between $y_1$ and $y_2$ with randomly chosen $s$ between $s_1$ and $s_2$, and construct the labeled dataset $\mathcal{D}_{t}=\{(s,x,y_l,y_w)|x \in X_{t}\}$.

In addition, to improve the effectiveness of online preference learning, we propose progressively increasing the training examples' difficulty akin to curriculum learning \citep{bengio2009curriculum}.
To this end, we simply reduce the temperature used for system prompt sampling as the iteration increases, which reduces the diversity between two system prompts. 
We expect that it also reduces the distance between two responses $y_1$ and $y_2$ from online response sampling with $\pi_{t-1}$ and ${S}_{t}$, \textit{i.e.}, more difficult to learn; therefore, this approach improves the effectiveness of online preference learning by continuously increasing the difficulty of the task.  
We present full procedure of \name{} in Algorithm \ref{alg:main}.

\section{Experiments}

\subsection{Setups}\label{Experimental_setup}

In this section, we first present our experimental setups. 
As denoted in Sec.~\ref{sec:3}, we adopt SPA framework \citep{kim2025debiasing} as our online preference learning algorithm for the experiments.
SPA enables the effective alignment of LLMs with limited preference data and does not require the external reward model; SPA includes the process of using initial seed data to train and create the initial DPO model. 
Here, the initial DPO model acts as the base model as well as the reward model before the iterative learning process begins (Eq. \ref{eq:selfee_pref} and \ref{eq:selfee_label}).

\noindent\textbf{Models.} For the policy LLM, we utilize an open-source model supervised fine-tuned (SFT) on UltraChat data \citep{ding2023enhancing} based on the Mistral-7B-0.1v model \citep{jiang2023mistral}, following the Zephyr recipe \cite{tunstall2023zephyr}.
For the feature classifier $q_{\phi}$ (Sec. \ref{sec:4.2}), we employ DeBERTa-v3-large \citep{he2023debertav3} as the backbone. 
We create five separate classifiers, one for each class of preference feature.   

\noindent\textbf{Datasets.} 
For the initial labeled preference data, we use UltraFeedback dataset \citep{cui2023ultrafeedback} which has been extensively used by prior works \citep{snorkel2024pairrm, rosset2024direct, kim2025debiasing}. 
Specifically, we sample 10K samples to construct a seed dataset. 
For \name{}, the seed data would be taken feature extraction and system prompt synthesis processes, and the resulting data with added system prompts are used for initial DPO training and feature classifier training. 
Excluding seed data, we sample 4 datasets of 5K input prompts each, ensuring no overlap; these datasets are used to generate responses in each iteration of online learning.

\noindent\textbf{Baselines.} 
We consider various preference learning baselines: \textit{DPO} \citep{rafailov2023direct}, \textit{Iterative DPO} \citep{xiong2024iterative}, and \textit{SPA} \citep{kim2025debiasing}.
All models under different baselines are initialized with the same SFT model. 
Iterative DPO, SPA, and \name{} use the same online instruction datasets for each iteration. 
For the reward model in Iterative DPO, we employe PairRM \citep{jiang2023llm}, which is widely used in LLM alignment. 
While the initial DPO model was originally adopted as a base model only for \name{} and SPA, we also consider using initial DPO as a base model in the case of Iterative DPO for a fair comparison.
Specifically, we train initial DPO model using the seed dataset without mapped system prompts.

\noindent\textbf{Evaluations.} 
To evaluate trained models, we employ commonly used benchmarks in preference alignment research as follows. 
AlpacaEval 2.0 \citep{dubois2024length} is designed to approximately evaluate human preference for instruction following, and calculates the win rate by comparing the response of GPT-4 \citep{openai2023gpt4} and the target model response by using GPT-4 as the evaluator.
It is known that this benchmark reflects human preferences well, including a length-controlled win rate that reduces the impact of length bias. 
Next, MT-Bench \citep{zheng2023judging} is designed to evaluate more diverse capabilities of LLMs by utilizing GPT-4 to score the responses of the model under evaluation on a scale from 0 to 10. 
Lastly, we evaluate the performance at the Anthropic-HHH \citep{askell2021general} which is designed to evaluate the alignment of LLMs with respect to three key attributes: Helpfulness, Honestness (\textit{i.e.} accuracy), and Harmlessness. 
For each data consisting of a human query and two corresponding AI responses, we assess model alignment by measuring whether the internal preference (Eq.~\ref{eq:selfee_pref}) is aligned with human labeled preference.
In addition, to measure the debiasing effect on preference features, we extract the preference features from the responses generated for the test instructions in AlpacaEval 2.0. 
Then, we use GPT-4o \citep{openai2024gpt4o} to infer the most prominent preference feature in each response. 
After obtaining the feature distribution, we measure how the KL divergence between this and the feature distribution of the responses of the initial model, which is trained on seed dataset. 

\noindent\textbf{Implementation details.} 
We extract preference features of the seed prefernece data using GPT-4o.
Here, we set the temperature as 0 and employ zero-shot chain-of-thought (CoT) prompting \citep{wei2022chain, kojima2022large}. 
We train the feature classifiers using Adam optimizer \citep{kingma2015adam} with a learning rate of 1e-5, a batch size of 32, over 5 epochs. 
We also use GPT-4o to synthesize system prompts, taking preference features as input. 
For double prompt sampling and scheduling (Sec.~\ref{sec:4.3}), the system prompts in the first iteration are generated with a temperature of 1.25, decreasing by 0.25 with each subsequent iteration. 
If scheduling is not applied, system prompts are generated with a temperature of 1. 
For subsequent iterations and the initial DPO, we set $\beta=0.1$ and train for 1 epoch with a batch size of 32. 
This value is the same throughout \name{} and SPA learning, but in the case of Iterative DPO, $\beta=0.01$ was used during online learning. 
The learning rate of 5e-7 is used with AdamW optimizer \citep{loshchilov2017fixing}. 
We employ a cosine learning rate scheduler with a 0.1 warm-up ratio. 
For \name{}, Iterative DPO, and SPA, response sampling is performed twice per prompt with a temperature of 0.7. 
Unlike the original SPA, we remove self-refine step to reduce the number of hyperparameters. 
All the prompts used with GPT-4o are presented in Appendix \ref{Prompt_Set}. 

\subsection{Main results}

\begin{figure*}[h]
\vspace{-0.2in}
\begin{center}
    {
    \subfigure[Harmlessness]
        {
        \includegraphics[width=0.31\textwidth]{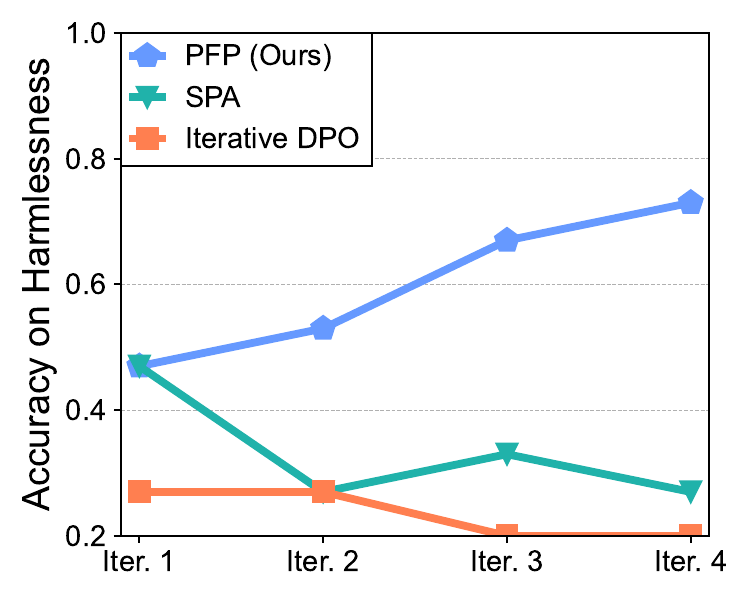}
        \label{fig:antro_harm}
        }
    \subfigure[Helpfulness]
        {
        \includegraphics[width=0.31\textwidth]{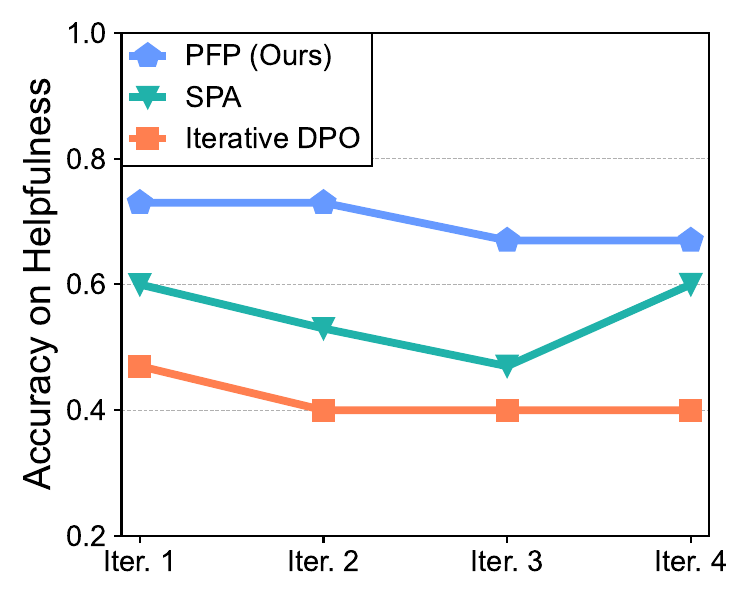}
        \label{fig:antro_help}
        }
    \subfigure[Honestness]
        {
        \includegraphics[width=0.31\textwidth]{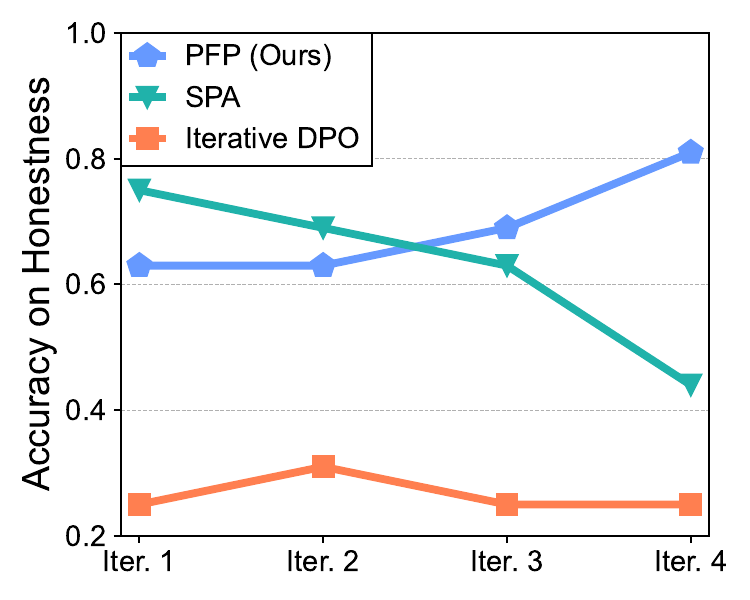}
        \label{fig:antro_hone}
        }
    }
\end{center}
\vspace{-0.15in}
\caption{\textbf{Change of Accuracy on Anthropic-HHH.} Accuracy of the model's internal preference (Eq.~\ref{eq:selfee_pref}) on Anthropic-HHH. For the reference model for Eq.~\ref{eq:selfee_pref}, the initial DPO model is used. 
As training progresses, \name{} exhibits significant improvements in harmlessness and honestness, whereas other baselines show notable declines.}  
\label{fig:antrophic_hhh}
\vspace{-0.1in}
\end{figure*}

\begin{table}[t!]
\centering
\caption{Results with different variants of Mistral-7B-v0.1. The best scores are highlighted in \textbf{bold}.}
\begin{adjustbox}{width=1.0\columnwidth}
\begin{tabular}{l|cc|c}
\toprule
\multirow{3}{*}{Models}& \multicolumn{2}{c|}{\textbf{AlpacaEval 2.0}} & \textbf{MT-Bench} \\
\cmidrule(lr){2-3} \cmidrule(lr){4-4}
 & \begin{tabular}{@{}c@{}}Len-control. \\ Win Rate  (\%)\end{tabular}  & \begin{tabular}{@{}c@{}}Avg. len \\ (\# chars)\end{tabular} & \begin{tabular}{@{}c@{}}Avg. Score \\ (0-10)  \end{tabular} \\ \midrule
SFT  & 7.58 & {901} & 6.34 \\
DPO (W/o sys) &  9.93  & 1409 & 6.34 \\
DPO (W sys) &  9.27 & 1135 & 6.61 \\
\midrule
SPA  & 14.23 & 2412 & 6.56  \\ 
Iterative DPO  & 13.13 & 1709 & 6.53  \\ 
\name{} (Ours) & \textbf{15.24}  & {1187} & \textbf{6.88} \\
\bottomrule
\end{tabular}
\end{adjustbox}
\vspace{-0.1in}
\label{tab:main_result}
\end{table}

In Table~\ref{tab:main_result}, we present the performance of the models obtained after 4 iterations of online preference learning, including the results of baselines such as used SFT and DPO trained on initial preference dataset. 
Notably, \name{} achieves the higher performance than SPA (7.58 $\rightarrow$ 14.23) and Iterative DPO (7.58 $\rightarrow$ 13.13) with a performance improvement of (7.58 $\rightarrow$ 15.24) based on AlpacaEval 2.0 length-controlled win rate. 
In MT-Bench, \name{} also showed a large improvement (6.34 $\rightarrow$ 6.88) compared to SPA (6.34 $\rightarrow$ 6.56) and Iterative DPO (6.34 $\rightarrow$ 6.53). 
These significant improvements in both AlpacaEval 2.0 or MT-Bench clearly affirm the overall enhancement in performance from \name{}. 

Next, in Fig.~\ref{fig:antrophic_hhh}, we present the performance trajectory for each method on the Anthropic-HHH benchmark.
In particular, \name{} consistently maintains high performance in all categories evaluated during the learning process; \name{} shows steady improvements in both harmlessness ($0.47 \to 0.73$) and honestness ($0.63 \to 0.81$). 
In contrast, Iterative DPO and SPA exhibit significant performance declines in these two metrics over the iterative training process. 
To verify whether these gains from \name{} stems from preserving the initial preference distribution and mitigating the bias, We compare the KL divergence of the responses from each method and the preference distribution from the initial DPO model (see more details in Eq.~\ref{eq:kl_equation}).  
The results are presented in Fig.~\ref{fig:bias}. 
In the case of Iterative DPO and SPA, the distribution diverges at the end, while in the case of \name{}, the marginal change in distribution occurs as iteration progresses. 
This represents that the existing iterative improvement algorithm has bias at the feature level, and \name{} sufficiently alleviates this. 
Additional experimental results including the effectiveness on LLaMA3-8B \citep{dubey2024llama} are presented in Appendix \ref{supp:more_results}.

\begin{figure*}[t]
\vspace{-0.2in}
\begin{center}
    {
    \subfigure[KL divergence for Table \ref{tab:ablation_labeling}]
        {
        \includegraphics[width=0.31\textwidth]{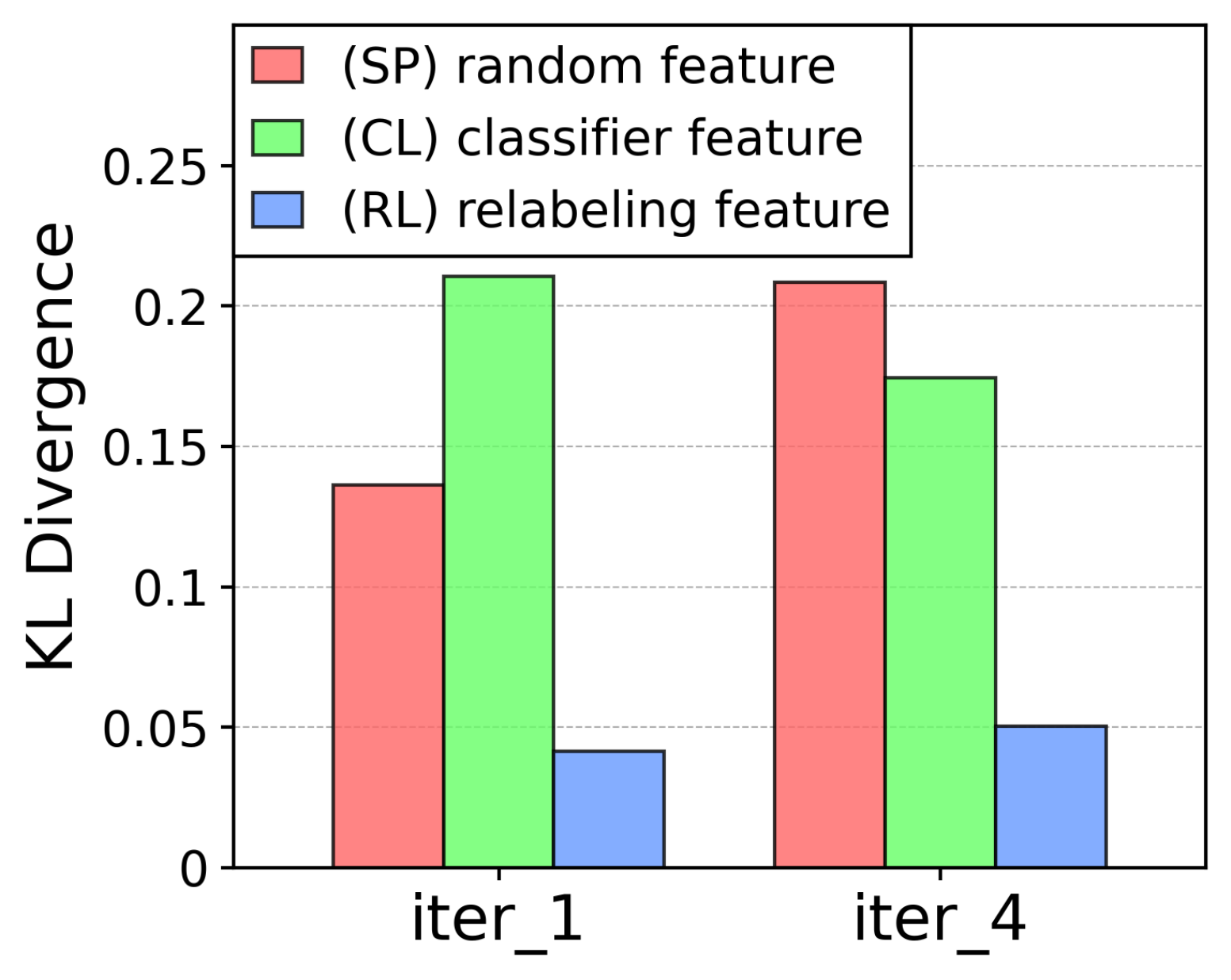}
        \label{fig:labeling_ablation}
        }
    \subfigure[KL divergence for Table \ref{tab:ablation_scheduling}]
        {
        \includegraphics[width=0.31\textwidth]{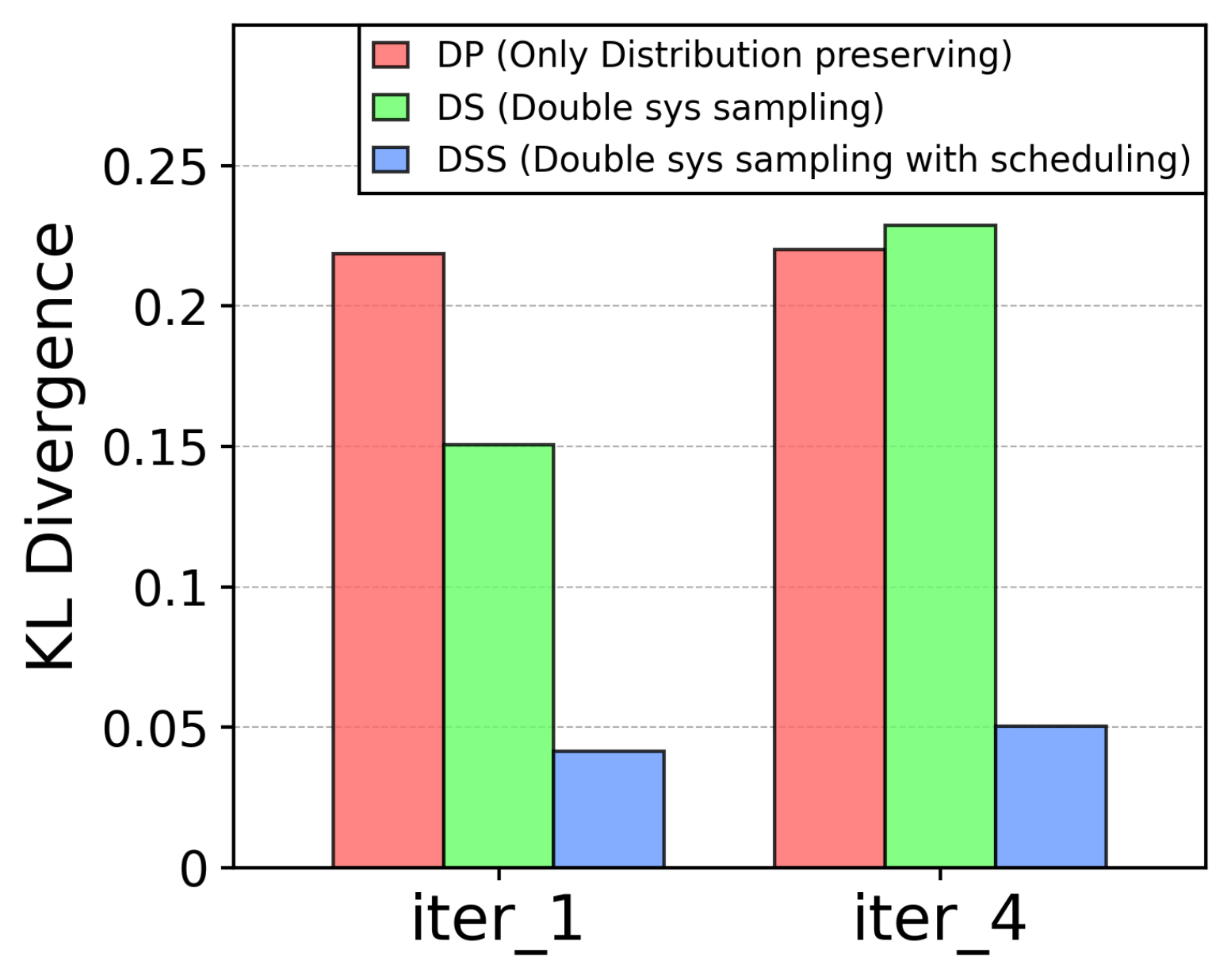}
        \label{fig:scheduling_ablation}
        } 
    }
    \subfigure[Effect on length bias]
        {
        \includegraphics[width=0.31\textwidth]{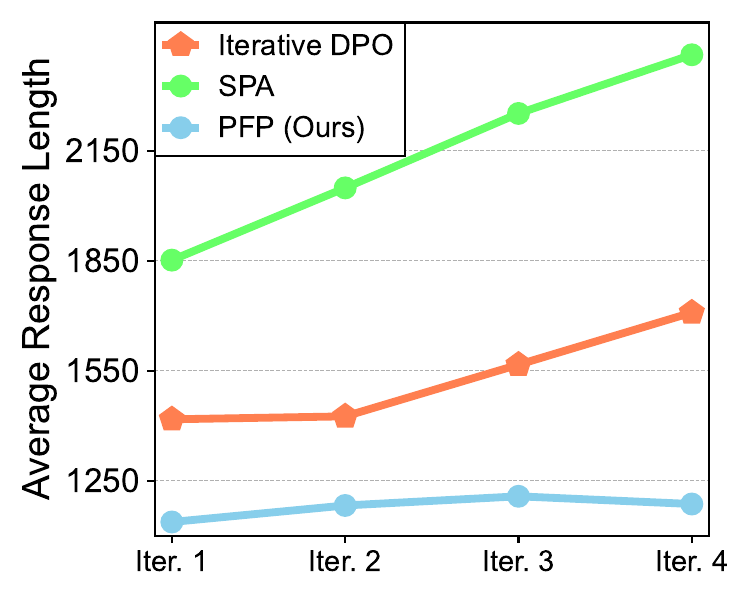}
        \label{fig:length_more}
        }
\end{center}
\vspace{-0.15in}
\caption{{\textbf{Analyses.} (a,b) KL divergence with feature distribution from different methods for ablation, (c) Change of average response length on AlpacaEval 2.0 with different methods.}
}
\label{fig:app_motivation}
\vspace{-0.1in}
\end{figure*}

\subsection{Ablation study}
To evaluate the effect of the feature labeling method (Sec.~\ref{sec:4.2}), we conduct an ablation study by removing some components. 
Table~\ref{tab:ablation_labeling} shows the corresponding experimental results after a total of 4 iterations.
For the method in 1st row, the random feature is created by generating a preference feature regardless of the prompt.
For the others, the preference features are sampled based on the probability of the feature classifier for each input instruction.
Additionally, for the 3rd row, we relabel the probability of the features according to Eq.~\ref{eq:opt} to preserve the original human preference distribution. 
In the results, it is observe that the sampled feature through the classifier is solely not sufficient; the performance is increased (11.58 $\rightarrow$ 12.52) on AlpacaEval 2.0. 
However, on MT-bench, the performance is decreased (6.76 $\rightarrow$ 6.45). 
However, by applying relabeling technique, the performances on both benchmarks are successfully improved.  
Meanwhile, as shown in Fig.~\ref{fig:labeling_ablation}, the preference feature distribution is successfully preserved when applying both feature classifier and relabeling.

\begin{table}[t!]
\centering

\caption{Results with iteratively trained models (from initial DPO) under different configurations of \name{}. SP, CL, RL are abbreviations of system prompt, classifier label, and relabeling, respectively. When using only the system prompt, features are mapped randomly.} 
\begin{adjustbox}{width=1.0\columnwidth}
\begin{tabular}{ccc|cc|c}
  \toprule
  \multicolumn{3}{c|}{\textbf{Components}} & \multicolumn{2}{c|}{\textbf{AlpacaEval 2.0}} & \textbf{MT-Bench} \\ 
  \cmidrule(lr){1-3} \cmidrule(lr){4-5} \cmidrule(lr){6-6}
   SP & CL & RL & \begin{tabular}{@{}c@{}}Len-control. \\ Win Rate (\%)\end{tabular} & \begin{tabular}{@{}c@{}}Avg. len \\ (\# chars)\end{tabular} & \begin{tabular}{@{}c@{}}Avg. Score \\ (0-10) \end{tabular} \\  \midrule
    \cmark & \xmark & \xmark &  11.58   & 1211 & 6.76  \\
\cmark & \cmark & \xmark & 12.52 & 1226 & 6.45 \\ 
  \cmark & \cmark & \cmark  & {15.24} & {1187} & {6.88}  \\ 
\bottomrule
\end{tabular}
\end{adjustbox}
\vspace{-0.1in}
\label{tab:ablation_labeling}
\end{table}
\begin{table}[t!]
\centering
\caption{Results with iteratively trained models (from initial DPO) under different configurations of \name{}. DP, DS, DSS are abbreviations of distribution preserving, double system prompt sampling, and double system prompt sampling with scheduling, respectively. } 
\begin{adjustbox}{width=1.0\linewidth}
\begin{tabular}{ccc|cc|c}
  \toprule
  \multicolumn{3}{c|}{\textbf{Components}} & \multicolumn{2}{c|}{\textbf{AlpacaEval 2.0}} & \textbf{MT-Bench} \\ 
  \cmidrule(lr){1-3} \cmidrule(lr){4-5} \cmidrule(lr){6-6}
   DP & DS & DSS & \begin{tabular}{@{}c@{}}Len-control. \\ Win Rate (\%)\end{tabular} & \begin{tabular}{@{}c@{}}Avg. len \\ (\# chars)\end{tabular} & \begin{tabular}{@{}c@{}}Avg. Score \\ (0-10)\end{tabular} \\  \midrule
    \cmark & \xmark & \xmark & 10.79 & 1276 & 6.69  \\
\cmark & \cmark & \xmark & 12.30  & 1217 & 6.56 \\ 
  \cmark & \cmark & \cmark  & {15.24} & {1187} & {6.88}  \\ 
\bottomrule
\end{tabular}
\end{adjustbox}
\vspace{-0.1in}
\label{tab:ablation_scheduling}
\end{table}

In addition, to evaluate the effect of the response sampling method (Sec.~\ref{sec:4.3}), we conduct additional ablation study by varying the usage of double system prompt sampling and scheduling. 
As shown in Table~\ref{tab:ablation_scheduling}, the double system prompt sampling is effective to improve the performance on AlpacaEval 2.0.
When scheduling is further applied, the improvement is enlarged, with AlpacaEval 2.0 (12.30 $\rightarrow$ 15.24) and MT-Bench (6.56 $\rightarrow$ 6.88). 
Additionally, these components not only improve performance but also play a significant role in bias mitigation. 
As shown in Fig.~\ref{fig:scheduling_ablation}, double system prompt sampling and scheduling are greatly effective to preserve the feature distribution. 
These results confirm that the proposed double system prompt sampling and scheduling are key factors to enhance performance by mitigating the bias.

\subsection{Mitigation of Length Bias} 
From the previous experiments, we observe that \name{} exhibits an unexpected additional advantages to mitigate \textit{length bias} of LLM (see Fig.~\ref{fig:length_more}), where aligned LLMs tend to generate and favor the longer responses \citep{park2024disentangling, singhal2023long}.  
As it becomes critical problem by complicating the accurate assessment of LLM performance \citep{dubois2024length, wang2023far}, many works have proposed to tackle this problem.
For example, the length penalty method works by heuristically subtracting a bias based on the length in the reward term  \citep{dong2024rlhf}.
Alternatively, R-DPO approach \citep{park2024disentangling} adds additional length regularization into DPO objective. 
Here, the common point is that the difference in length between two sentences is simply processed heuristically. 
To evaluate the effectivness of \name{} to mitigate length bias, we compare \name{} with the length penalty method and R-DPO applied to Iterative DPO. 
We have tried both methods, and R-DPO method with $\alpha=0.01$ exhibits the best performance.
However, as shown in Table~\ref{tab:length_penality}, the overall reduction in length remained limited and \name{} is more effective in reducing length. 
This result again shows the importance of preventing preference bias during online learning. 
More details about these experiments are presented in Appendix \ref{lc_control}. 

\begin{table}[t!]
\centering
\caption{Results with iteratively trained models (from initial DPO) under different methods to mitigate length bias (length penalty and R-DPO). The best scores are highlighted in \textbf{bold}.} 
\begin{adjustbox}{width=1.0\columnwidth}
\begin{tabular}{l|cc|c}
  \toprule
  \multirow{3}{*}{Methods}& \multicolumn{2}{c|}{\textbf{AlpacaEval 2.0}} & \textbf{MT-Bench} \\ 
  \cmidrule(lr){2-3} \cmidrule(lr){4-4}
   & \begin{tabular}{@{}c@{}}Len-control. \\ Win Rate (\%)\end{tabular} & \begin{tabular}{@{}c@{}}Avg. len \\ (\# chars)\end{tabular} & \begin{tabular}{@{}c@{}}Avg. Score \\ (0-10) \end{tabular} \\ 
   \midrule
Iterative DPO & 13.13 & 1709 & 6.53 \\
w\textbackslash~ length penalty & 12.19 & 1689 & 6.60 \\ 
w\textbackslash~ R-DPO & 13.07 & 1613 & 6.80 \\ \midrule
\name{} (Ours) & \textbf{15.24} & \textbf{1187} & \textbf{6.88} \\ 
\bottomrule
\end{tabular}
\end{adjustbox}
\label{tab:length_penality}
\vspace{-0.1in}
\end{table}
\section{Conclusion} 
We propose \name{}, a novel framework that explicitly preserves preference features during the online preference learning to mitigate potential bias. 
We demonstrate that incorporating preference features from human feedback into system prompts and preserving the feature distribution over each iteration of online learning effective in preventing bias. 
This approach not only aligns human preferences more effectively than the existing methods but also eliminates length bias and undesired preference feature biases, while uniformly improving various factors such as harmlessness and honestness.

\newpage
\section*{Acknowledgments}

This work was supported by Institute for Information \& communications Technology Promotion (IITP) grant funded by the Korea government (MSIT) (No.RS-2019-II190075 Artificial Intelligence Graduate School Program (KAIST); No. RS-2024-00509279, Global AI Frontier Lab and NIPA( National IT Industry Promotion Agency), through the Ministry of Science and ICT (Hyperscale AI flagship project).

\section*{Limitations}

Extracting preference features and generating system prompts currently requires powerful LLMs like GPT-4o \citep{openai2024gpt4o}, which requires additional computational costs. 
Future work should explore the use of smaller LLMs such as LLaMA-3-8B \citep{dubey2024llama} for this process. Additionally, further research is needed to assess the impact of incorporating system prompts into the supervised fine-tuning (SFT) stage of training. 

\section*{Broader impact and ethical implications}

We anticipate that \name{} will significantly advance the development of real-world AI systems by mitigating a range of biases and promoting safer models. 
Our approach leverages standard preference alignment datasets to address biases associated with key safety attributes such as harmlessness and honestness, while also reducing issues like length bias and preference feature-level bias. 
By effectively curbing these unintended biases, \name{} enables the creation of AI models that not only achieve state-of-the-art performance but also adhere to high ethical standards.
However, because the effectiveness of debiasing by \name{} relies on a predefined preference feature set, all biases cannot be completely removed, and hence completely trusting the model trained in this way can be potentially dangerous.
Consequently, this work can contribute the deployment of AI applications that are both effective and socially responsible.

\bibliography{acl2025_conference}

\newpage
\appendix

\section{More Details}\label{model_details}
This section provides more details about the experimental setups in Section 4. We note that all of our experiments are conducted with 4 NVIDIA RTX 854 A6000 GPUs (48GB memory) and AMD EPYC 7313 16-core Processor (3.7 max CPU Ghz).
Training models with each online preference learning method (Iterative DPO, SPA, and \name{}) takes approximately 1 to 2 days.

\noindent\textbf{Model and Dataset.} 
In our experiments, we do not perform supervised fine-tuning separately. 
Instead, we utilize open-source models trained according to the Zephyr recipe. 
Specifically, we employ the models \texttt{zephyr-7b-sft-full}\footnote{\url{https://huggingface.co/alignment-handbook/zephyr-7b-sft-full}} and \texttt{Llama-3-Base-8B-SFT}\footnote{\url{https://huggingface.co/princeton-nlp/Llama-3-Base-8B-SFT}}. 
These models are based on the \texttt{mistral-7b-0.1}\footnote{\url{https://huggingface.co/mistralai/Mistral-7B-v0.1}} and \texttt{llama-3-8b}\footnote{\url{https://huggingface.co/meta-llama/Meta-Llama-3-8B}} base models, respectively, and were trained using the \texttt{ultrachat\_200k}\footnote{\url{https://huggingface.co/datasets/HuggingFaceH4/ultrachat_200k}} dataset. 
For training SPA and \name{}, we use the \texttt{ultrafeedback} preference dataset\footnote{\url{https://huggingface.co/datasets/argilla/ultrafeedback-binarized-preferences-cleaned}}.

\noindent\textbf{Evaluation.} 
Here, we present more details how to measure KL divergence between responses of the trained models and the initial distribution in seed dataset. 
First, we extract the preference features from the responses generated for the test instructions in AlpacaEval 2.0. 
Next, we use GPT-4o \citep{openai2024gpt4o} to infer the most prominent preference feature in each response. 
Then, we measure the KL divergence between this and the feature distribution of the responses of the initial model as follows: 
\begin{align}\label{eq:kl_equation}
   & D_{\tt KL}(P_{\tt Init. Model} \parallel P_{\tt target}) \nonumber \\ &= \sum_{x} P_{\tt Init.Model}(x) \log \left( \frac{P_{\tt Init. Model}(x)}{P_{\tt target}(x)} \right).
\end{align}
where initial (Init.) model is corresponding to SFT when measuring KL for initial DPO, and initial DPO for others.

\section{Pre-defined Preference Feature Set}\label{Peature_Set} 
Table~\ref{tab:featureset} shows the pre-defined preference feature set $\mathcal{P}$. 
The definition of the preference feature set was referenced from Janus \cite{lee2024aligning}. 
Preference features consist of 5 different classes (i.e. Style, Tone, etc), and each class gets 5 different sub-features (i.e. Clarity, Conciseness, etc). Each preference is defined by a total of five sub-features, with one sub-feature assigned per class.  

\begin{table}[t!]
    \centering
    \small
    \renewcommand{\arraystretch}{1.2}
    \caption{Predefined preference feature set.}
    \resizebox{\linewidth}{!}{
    \begin{tabular}{c |c}
        \toprule 
        \textbf{Domain} & \textbf{Feature Set} \\
        \midrule 
        \multirow{2}{*}{Style} & Clarity, Conciseness, \\
        & Format, Vividness, Consistency \\
        \midrule 
        \multirow{2}{*}{Tone} & Formal, Authoritative, \\
        & Sophisticated, Engaging, Familiar \\
        \midrule 
        \multirow{2}{*}{Harmlessness} & Sensitivity, Safety, \\
        & Accuracy, Morality, Trustworthiness \\
        \midrule 
        \multirow{2}{*}{User's Background Knowledge} & Basic, Novice, \\
        & Intermediate, Advanced, Expert \\
        \midrule 
        \multirow{2}{*}{Informativeness} & Relevance, Practicality, \\
        & Depth, Creativity, Efficiency \\
        \bottomrule
    \end{tabular}}
    \label{tab:featureset}
\end{table}

\section{Initial DPO Training Performance}\label{initial_dpo_performance}

\begin{figure}[t!]
\centering
\begin{adjustbox}{width=0.8\columnwidth}
   \includegraphics[width=\textwidth]{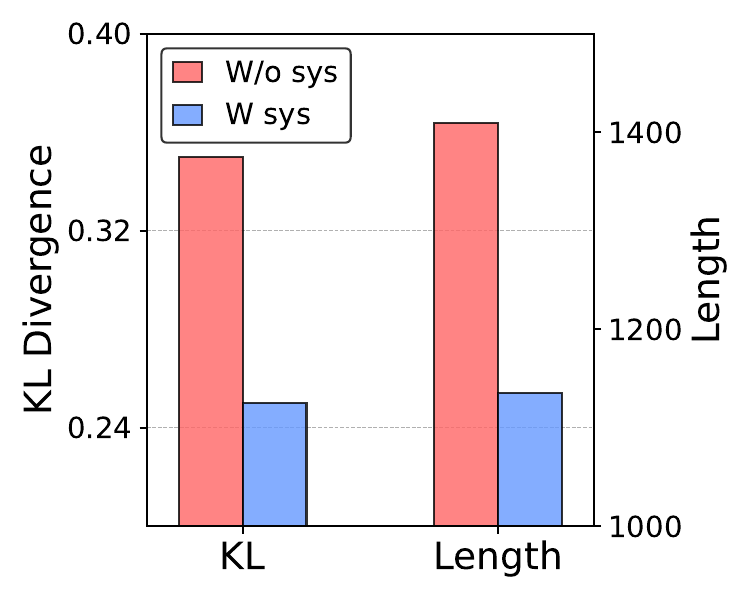}
\end{adjustbox}
\caption{\textbf{Initial DPO Analysis I.} LLMs trained by DPO using human feedback data with system prompt has less length
and feature distribution bias.}
\label{fig:inital_dpo_kl}
\end{figure}
\begin{figure}[t!]
\centering
\begin{adjustbox}{width=0.8\columnwidth}
   \includegraphics[width=\textwidth]{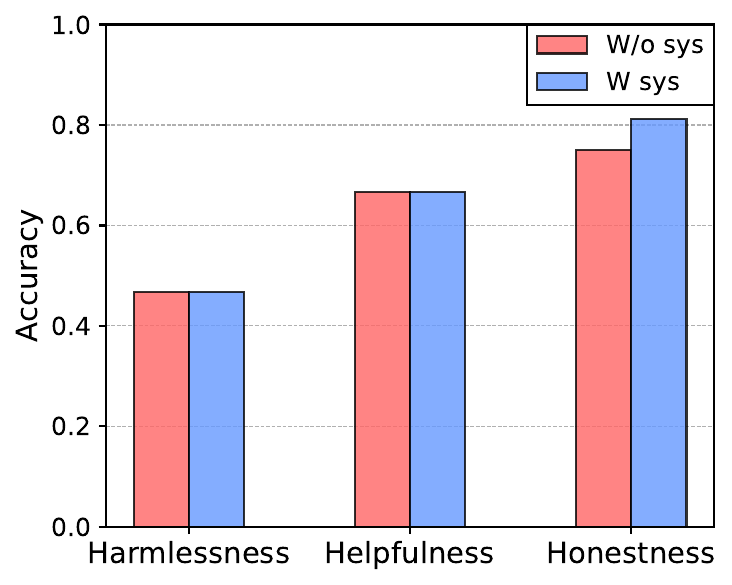}
\end{adjustbox}
\caption{\textbf{Initial DPO Analysis II.} LLMs trained by DPO using human feedback data with a system prompt have higher performance in honestness on Anthropic-HHH}
\label{fig:inital_dpo_hhh}
\end{figure}

In this section, we conduct the experiments to verify the effectiveness of using system prompt to handle preference feature. 
Specifically, we compare a DPO model trained with the preference feature from human feedback data explicitly included in the system prompt, against a model trained without feature. 
Based on AlpacaEval 2.0, the model trained with the system prompt performs slightly worse (9.93 vs 9.27), but based on MT-bench, a model trained with the system prompt gets a higher score than others (6.34 vs 6.61) (see Table~\ref{tab:main_result}). 
However, as shown in Fig.~\ref{fig:inital_dpo_kl}, 
which KL divergence is measured from the SFT response distribution, the DPO model with the preference feature exhibits significantly reduced preference feature bias, and the length bias is also considerably decreased. 
These results suggest that explicitly considering the preference feature from human feedback data into the system prompt significantly aids in debiasing the model. 
Fig.~\ref{fig:inital_dpo_hhh} illustrates how each DPO model aligns with the SFT model on Anthropic-HHH. Overall, the alignment patterns are similar; however, the initial DPO model with the system prompt achieves marginally better performance in honestness.

\section{Baselines to Reduce Length Bias during Alignment}\label{lc_control}

\textbf{Length penalty.} 
We applied the length penalty according to the RLHFlow approach \citep{dong2024rlhf}. 
This is a method to apply a length penalty at the labeling stage by adjusting the reward of the reward model according to Eq. \ref{eq:lc_penalty_eqation}. 
To find the efficient hyper-parameter for this baseline, we experimented with $\alpha = 0.01$, $0.001$, and $0.0001$ for iteration 1. 
Then, we applied the hyper-parameter that most effectively reduced length ($\alpha = 0.001$, see 3rd-5th rows in Table~ \ref{tab:length_penality_appendix_iter_1}) through iteration 4. 
As shown in Table~\ref{tab:length_penality_appendix_iter_1}, this approach often fails. 
Although $\alpha = 0.001$ showed the best reduction in length in iteration 1, the overall reduction in length remained limited and the performance was degraded as a result. 
This was the same even when iteration was extended. 
\begin{equation}\label{eq:lc_penalty_eqation}
r_{\tt penalty}(x,y) = r(x,y) - \alpha |y|
\end{equation}
\textbf{R-DPO.} For conduct R-DPO \citep{park2024disentangling}, we change DPO objective function to following Eq. \ref{eq:r-dpo}. 
Similar to the length penalty method, we experimented with $\gamma = 0.1$, $0.01$ for iteration 1, to find the effective hyper-parameter $\gamma$. 
We applied the hyper-parameter $\gamma = 0.01$ that effectively reduced length (see 6rd-7th rows in Table~\ref{tab:length_penality_appendix_iter_1}) through iteration 4. As observed in Table~\ref{tab:length_penality}, R-DPO successfully reduces the responses' length (1709 $\rightarrow$ 1613), but the reduction is still limited to resolve the length bias. 
These results show that heuristic length control is often unstable and does not work effectively.

\begin{equation}\label{eq:r-dpo}
    \small
    \setlength{\abovedisplayskip}{3pt}
    \setlength{\belowdisplayskip}{3pt} 
    \begin{aligned}
        \mathcal{L}_\text{\tt R-DPO}(\pi_{\theta}) &= 
        -\mathbb{E}_{(x, y_w, y_l) \sim \mathcal{D}} \Bigg[ \\
        &\quad \sigma \Bigg(
        \beta \log \frac{\pi_{\theta}(y_w|x)}{\pi_{\text{\tt ref}}(y_w|x)} - \beta \log \frac{\pi_{\theta}(y_l|x)}{\pi_{\text{\tt ref}}(y_l|x)}
        \Bigg) \\
        &\quad + \gamma (|y_{w}| - |y_{l}|)
        \Bigg]
    \end{aligned}
\end{equation}

\begin{algorithm}[t!]
   \caption{\name{} algorithm}
   \label{alg:main}
\begin{algorithmic}
  \State
  \textbf{Input:} initial LLM $\pi_{\text{init}}$, human preference dataset $\mathcal{D}$, number of online learning iterations $T$, new instruction sets $\{X_{t}\}_{t=1}^{T}$, feature extractor $\text{LLM}_{\text{FE}}$, system prompt synthesizer $\text{LLM}_{\text{SS}}$ 
  \vspace{0.01in} 
  \hrule
  \vspace{0.05in}  
  \State Extract preference features of $\mathcal{D}$ using $\text{LLM}_{\text{FE}}$ and construct $\mathcal{D}_{\text{FE}}$ (Sec. \ref{sec:4.1})
  \State Train feature classifier $q_{\phi}$ using $\mathcal{D}_{\text{FE}}$ (Sec. \ref{sec:4.2})
  \State $\pi_{0} \leftarrow \text{DPO}(\pi_{\text{init}}, \pi_{\text{init}}, \mathcal{D}_{\text{FE}})$ 
  \For{$t=1$ {\bfseries to} $T$}
  \State Assign preference features for $x \in X_{t}$ using $q_{\phi}$ and solving Eq. \ref{eq:opt}, and construct $\widetilde{X}_{t}$
  \State Sample two system prompts $s_{1}, s_{2}$ for $p \in \widetilde{X}_{t}$ using $\text{LLM}_{\text{SS}}$, and construct ${S}_{t}$
  \State Synthesize preference data $\mathcal{D}_{t}$ with $\pi_{t-1} \text{ and } {S}_{t}$ (Eq. \ref{eq:selfee_pref} and \ref{eq:selfee_label})
  \State $\pi_{t} \leftarrow \text{DPO}(\pi_{t-1}, \pi_{t-1}, \mathcal{D}_{\text{t}})$ 
  \EndFor
  \State \textbf{return}~~$\pi_{T}$
\end{algorithmic}
\end{algorithm}
\begin{table}[t!]
\centering
\small
\renewcommand{\arraystretch}{0.85}
\setlength{\tabcolsep}{3pt}
\caption{Results on AlpacaEval 2.0 under different methods to mitigate length bias (length penalty and R-DPO), with all comparisons corresponding to Iteration 1 results.
The best scores are highlighted in \textbf{bold}.
}
\vspace{0.05in}
\resizebox{\linewidth}{!}{
\begin{tabular}{l|cc}
  \toprule
  \multirow{3}{*}{Methods} & \multicolumn{2}{c}{\textbf{AlpacaEval 2.0}} \\ 
  \cmidrule(lr){2-3}
   & \begin{tabular}{@{}c@{}}LC \\ Win Rate\end{tabular} 
   & \begin{tabular}{@{}c@{}}Avg. len \\ (\# chars)\end{tabular} \\  
  \midrule
  Initial DPO &  9.93 & 1409 \\
  Iterative DPO (iter 1)  & 10.48 & 1418 \\
  \midrule
  w\textbackslash~ LP ($\alpha=10^{-2}$) & 11.02 & 1433 \\
  w\textbackslash~ LP ($\alpha=10^{-3}$) &  9.60 & 1406 \\
  w\textbackslash~ LP ($\alpha=10^{-4}$) & 10.72 & 1414 \\
  \midrule
  w\textbackslash~ R-DPO ($\gamma=10^{-1}$) & 9.99 & 1519 \\
  w\textbackslash~ R-DPO ($\gamma=10^{-2}$) & \textbf{11.09} & \textbf{1385} \\
  \bottomrule
\end{tabular}
}
\label{tab:length_penality_appendix_iter_1}
\end{table}

\section{Additional Results and Analyses}\label{supp:more_results} 

\begin{figure*}[t]
\begin{center}
    {
    \subfigure[Style]
        {
        \includegraphics[width=0.18\textwidth]{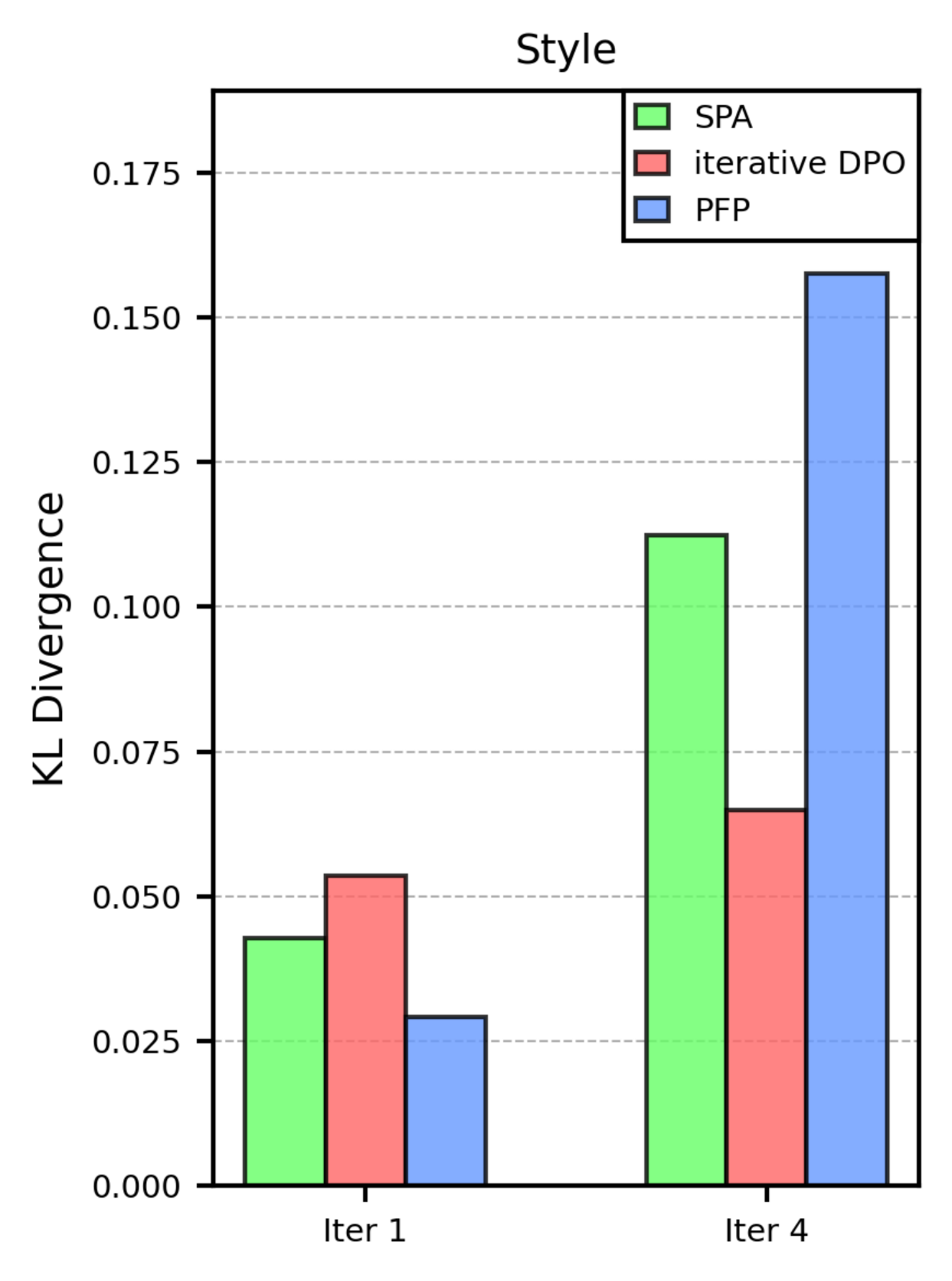}
        \label{fig:style}
        }
    \subfigure[Tone]
        {
        \includegraphics[width=0.18\textwidth]{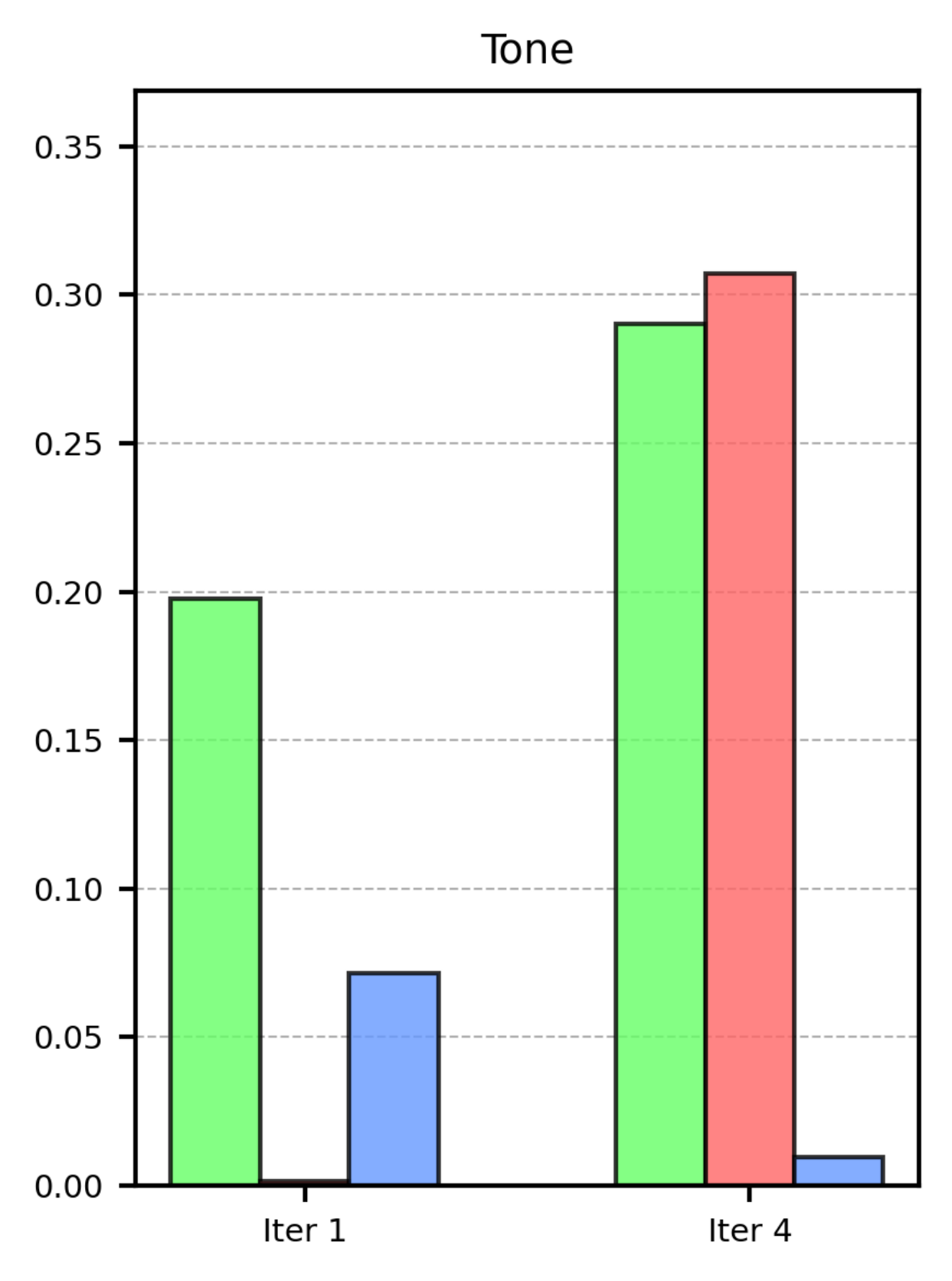}
        \label{fig:tone}
        }
    \subfigure[Harmless]
        {
        \includegraphics[width=0.18\textwidth]{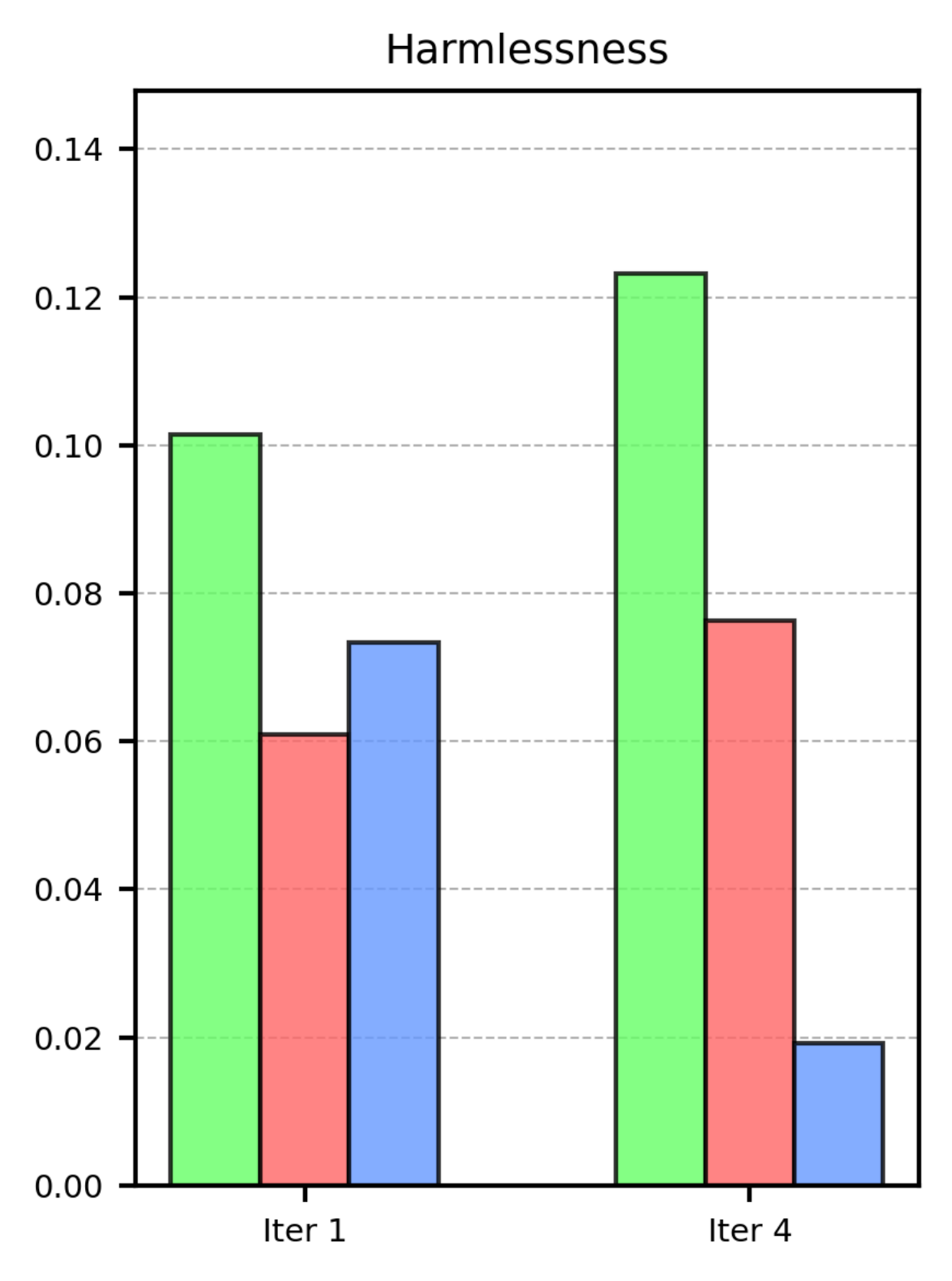}
        \label{fig:harmless}
        }
    \subfigure[Background]
        {
        \includegraphics[width=0.18\textwidth]{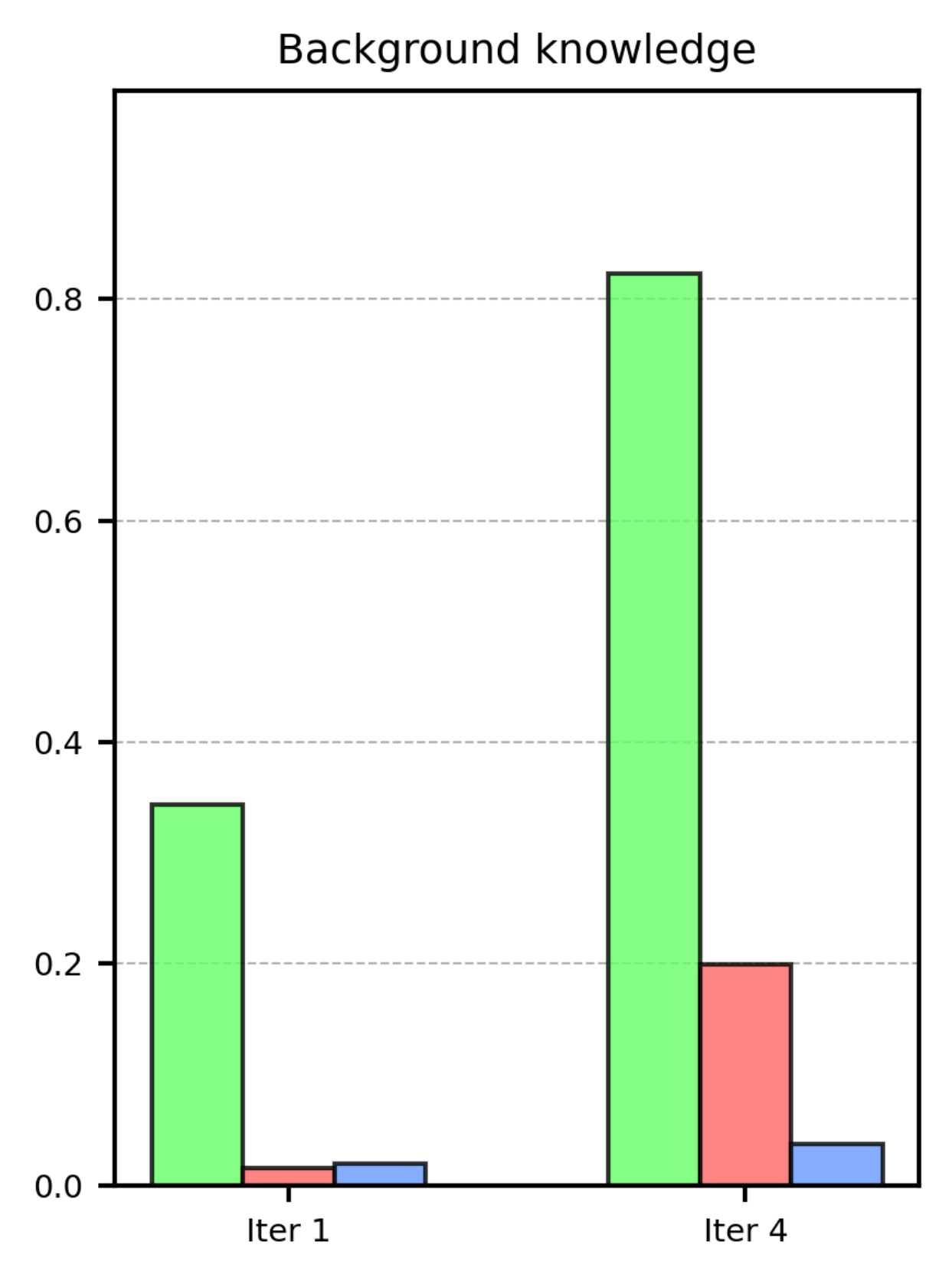}
        \label{fig:background}
        }
    \subfigure[Informative]
        {
        \includegraphics[width=0.18\textwidth]{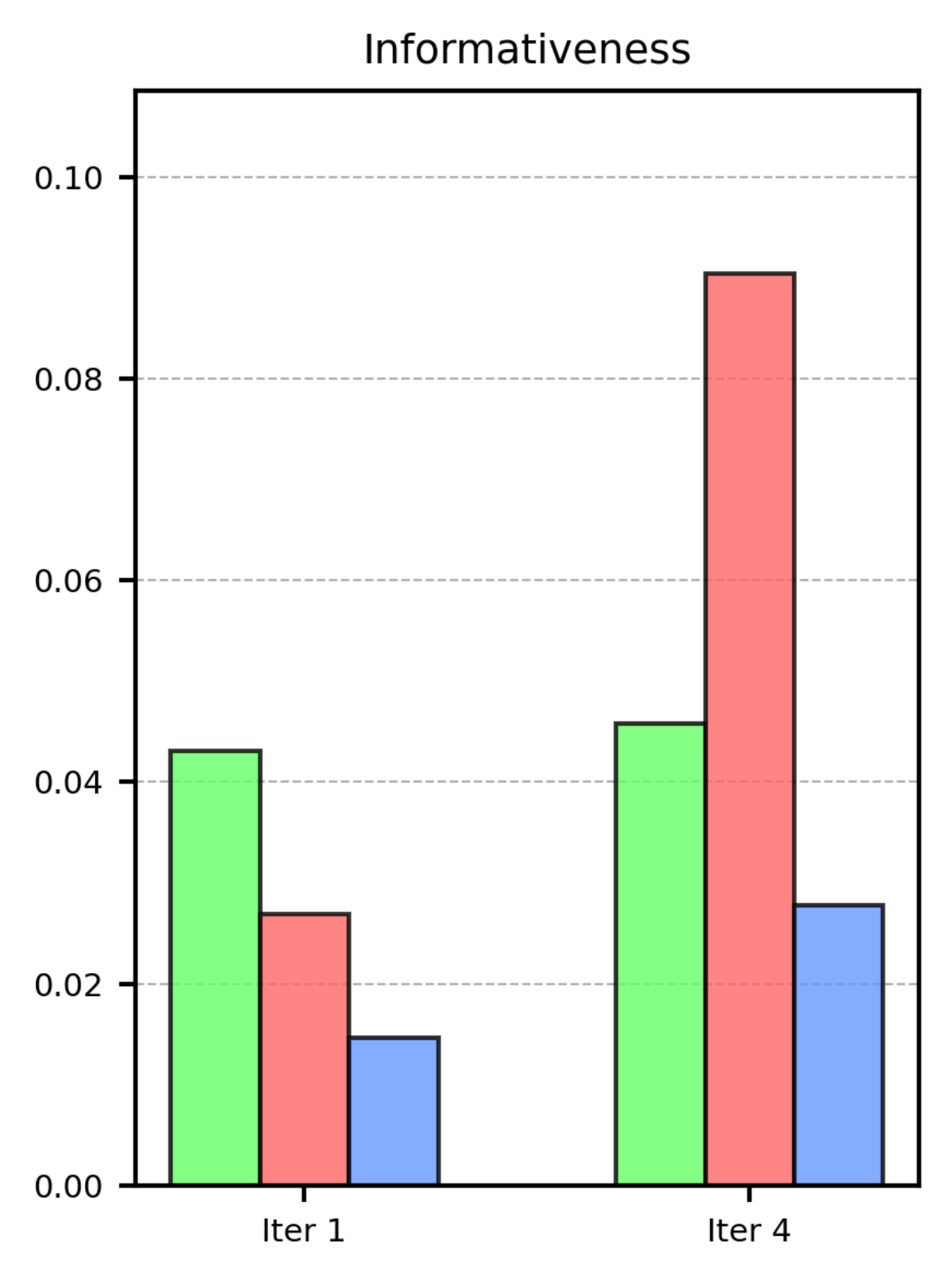}
        \label{fig:informative}
        }
    }
\end{center}
\vspace{-0.1in}
\caption{\textbf{Change of preference features.} KL divergence comparison by class, showing how the feature distribution of the initial DPO model's response evolves during the online learning process. \name{}, unlike other iterative learning algorithms, shows minimal change in distribution.}  
\label{fig:comparison}
\end{figure*}
\begin{table}[t!]
\centering
\caption{Results with different variants of LLaMA3.1-8B. The best scores are highlighted in \textbf{bold}.}
\begin{adjustbox}{width=1.0\columnwidth}
\begin{tabular}{l|cc|c}
\toprule
\multirow{3}{*}{Models}& \multicolumn{2}{c|}{\textbf{AlpacaEval 2.0}} & \textbf{MT-Bench} \\
\cmidrule(lr){2-3} \cmidrule(lr){4-4}
 & \begin{tabular}{@{}c@{}}Len-control. \\ Win Rate  (\%)\end{tabular}  & \begin{tabular}{@{}c@{}}Avg. len \\ (\# chars)\end{tabular} & \begin{tabular}{@{}c@{}}Avg. Score \\ (0-10)  \end{tabular} \\ \midrule
SFT  & 7.33 & {877} & 6.60 \\
DPO  &  10.12 & 1046 & 6.92  \\
\name{} (Ours) & \textbf{15.05} & {1082} & \textbf{7.08} \\
\bottomrule
\end{tabular}
\end{adjustbox}
\label{tab:main_result_llama}
\end{table}

\begin{figure*}[t]
\begin{center}
    \centering
    \includegraphics[width=\textwidth]{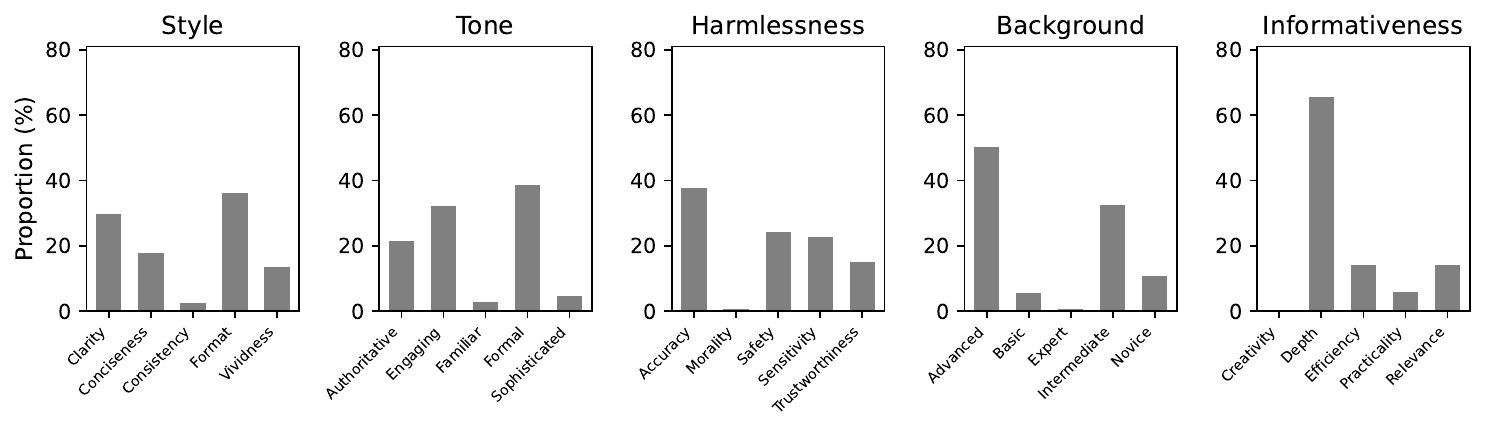}
    \label{fig:inital_feature_distribution}
\end{center}
\vspace{-0.1in}
\caption{\textbf{Preference feature distribution captured in seed dataset} }  
\label{fig:seed_distb}
\end{figure*}

\textbf{Accuracy of trained preference feature classifier.} 
In Table \ref{tab:acc_feature}, we additionally measure the test accuracy of the trained feature classifier on the separately constructed test dataset. 
Specifically, this test dataset is created by randomly selecting 917 samples from the initial dataset to be excluded from learning.
The results show moderate accuracy, which is limited by the small amount of training data and the long-tailed nature of preference features (see Figure \ref{fig:seed_distb}). 
These findings highlight the importance of the proposed distribution preservation step in addressing the limitations of the feature classifier.

\begin{table}[t!]
\centering
{
\footnotesize 
\renewcommand{\arraystretch}{0.85} 
\setlength{\tabcolsep}{3pt} 
\caption{ \textbf{Feature classifier accuracy.} Test accuracy of the trained feature classifier (in Section \ref{sec:4.2}) on the separately constructed test dataset.}
\resizebox{0.95\linewidth}{!}{
\begin{tabular}{l|ccccc}
\toprule
\text{Metric} & \makebox[1.8cm]{\text{background}} & \makebox[2cm]{\text{harmlessness}} & \makebox[2cm]{\text{informativeness}} & \makebox[1.4cm]{\text{style}} & \makebox[1.2cm]{\text{tone}}  \\ \midrule
\text{Accuracy} & 0.535 & 0.512 & 0.688 & 0.496 & 0.507 \\ 
\text{F1 Score} & 0.532 & 0.513 & 0.663 & 0.489 & 0.426 \\
\bottomrule
\end{tabular}
}
\label{tab:acc_feature}
}
\end{table}

\noindent\textbf{Robustness with potential variance.} 
Since our evaluation using AlpacaEval 2.0 rely on GPT-4 as the evaluator, this can potentially raise a question about the  variability in the results due to using GPT-4 for evaluation, particularly since its responses can introduce variance. 
To address this, we conducted two additional rounds of experiments. 
In these new experiments, we fixed the initial DPO model and repeated subsequent evaluations to assess consistency. 
We conducted an evaluation on AlpacaEval 2.0, and the results are presented in Table \ref{tab:rebuttal_variance}. 
While the differences in LC Win Rate may appear modest (\textit{e.g.}, 1.46\% improvement), the variance for these evaluations (\textit{e.g.}, 0.29 for \name{}) supports the statistical significance of these results. 
Moreover, we remark that the primary contribution of our method lies in feature debiasing. 
As shown in Figure \ref{fig:comparison}, \name{} demonstrates substantial improvements over other baselines in mitigating preference feature bias.

\begin{table}[t!]
\centering
\small
\renewcommand{\arraystretch}{1.2}
\caption{Evaluation results on AlpacaEval 2.0 with different random seeds.}
\resizebox{\linewidth}{!}{
\begin{tabular}{l|ccc|cc}
\toprule
Methods & 1st Seed & 2nd Seed & 3rd Seed & Average & Var \\ \midrule
\name{}: LC Win Rate (\%) & 15.24 & 14.38 & 14.22 & 14.61 & 0.29 \\
SPA: LC Win Rate (\%) & 14.23 & 12.58 & 12.64 & 13.15 & 0.84 \\
\bottomrule
\end{tabular}
}
\label{tab:rebuttal_variance}
\end{table}

\begin{figure*}[t]
\begin{center}
    \centering
    \includegraphics[width=\textwidth]{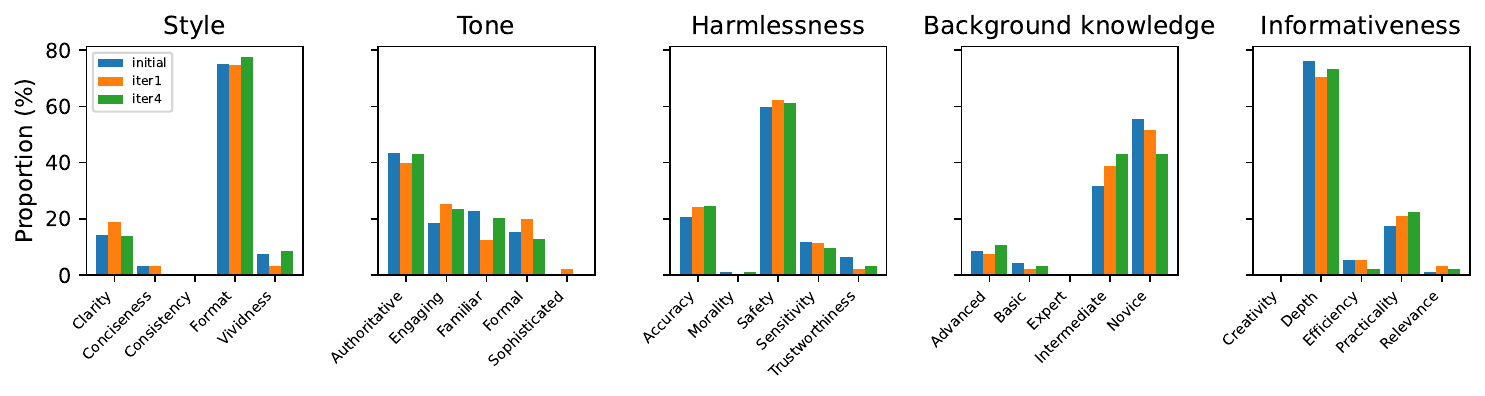}
\end{center}
\vspace{-0.1in}
\caption{\textbf{Preference feature distribution captured in responses generated from PFP} }  
\label{fig:PFP_distribution_change}
\end{figure*}

\begin{figure*}[t]
\begin{center}
    \centering
    \includegraphics[width=\textwidth]{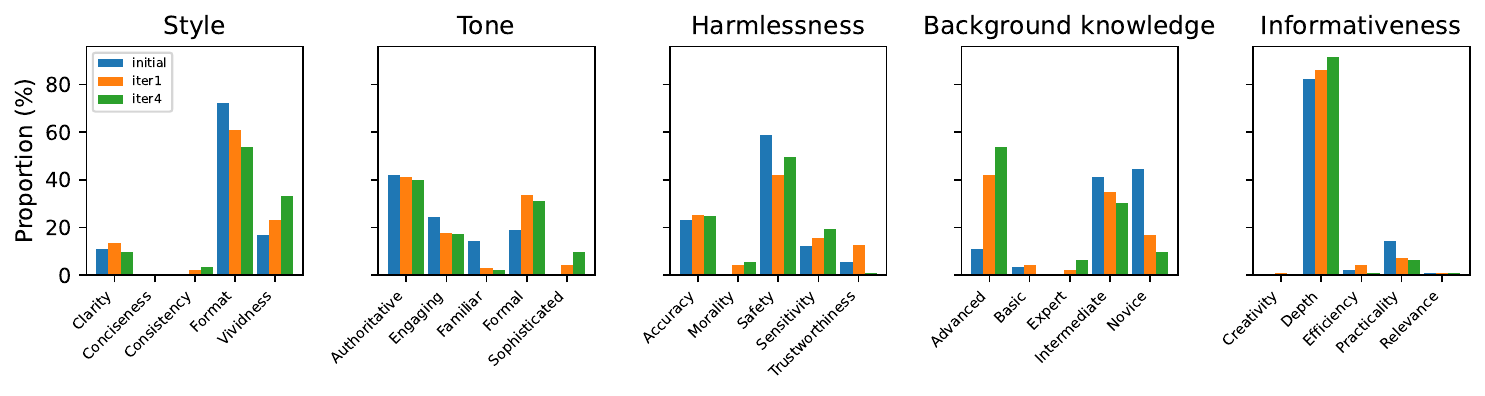}
\end{center}
\vspace{-0.1in}
\caption{\textbf{Preference feature distribution captured in responses generated from  SPA} }  
\label{fig:supp_selfee}
\end{figure*}

\begin{figure*}[t]
\begin{center}
    \centering
    \includegraphics[width=\textwidth]{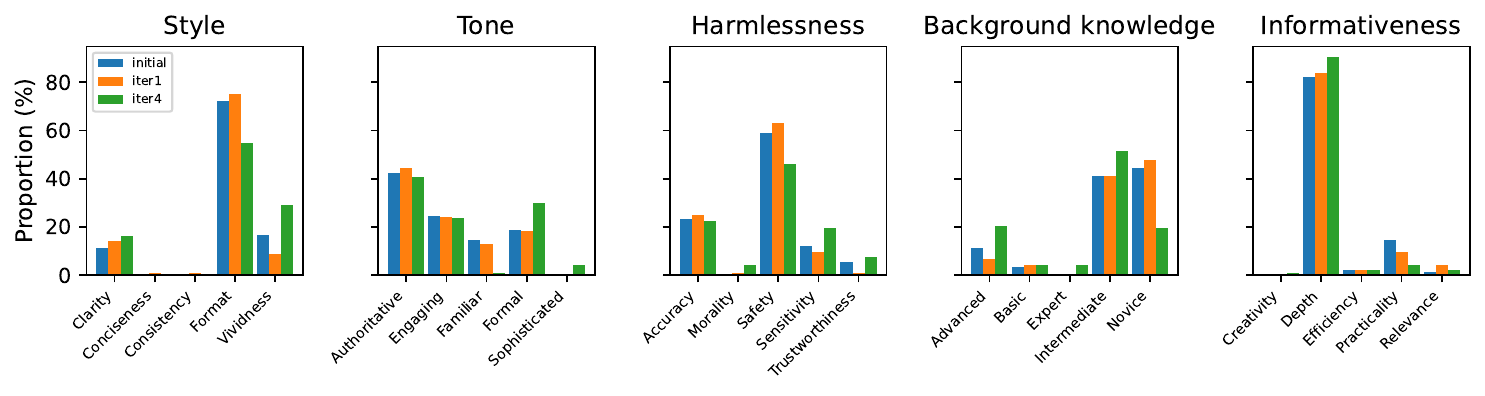}
\end{center}
\vspace{-0.1in}
\caption{\textbf{Preference feature distribution captured in responses generated from Iterative DPO} }  
\label{fig:iterative_dpo_distribution_change}
\end{figure*}

\noindent\textbf{Preference feature distribution. } 
Here, we present the preference feature distributions specifically. 
For each category of preference feature, we normalize the frequency and present the proportion of each sub-feature. 
Figure \ref{fig:seed_distb} is the distribution of seed preference dataset, which is extracted with feature extractor (see Section \ref{sec:4.1}). 
Remarkably, one can observe the imbalanced distribution for each category, which potentially affect to the classifier's performance. 
Next, in Figures \ref{fig:PFP_distribution_change}, \ref{fig:supp_selfee},
\ref{fig:iterative_dpo_distribution_change}, 
we present the preference feature distribution under different online preference learning methods. 
Unlike Figure \ref{fig:seed_distb}, this feature is measured by a single response generated from the AlpacaEval 2.0 prompt. 

\noindent\textbf{Experiment on difference dataset. } To investigate whether \name{} generalizes well to different datasets, we experiment with \name{} using different datasets. Specifically, we use  OpenHermesPreferences\footnote{\url{https://huggingface.co/datasets/argilla/OpenHermesPreferences}} to construct new prompt datasets (5K input prompts each) for iterative training of \name{}. With these datasets, we train Mistral-7B-0.1v with 2 iterations of \name{} training, using the same hyperparameters. As shown in Table \ref{tab:diverse_dataset}, \name{} consistently improves performance and outperforms the main baseline, SPA, especially on the Anthropic-HHH benchmark. This result confirms the robustness and generalizability of the proposed method.

\begin{table}[htbp]
\centering
\small
\renewcommand{\arraystretch}{1.2}
\caption{Evaluation results on various benchmark with OpenHermesPreferences dataset.} 
\resizebox{\linewidth}{!}{
\begin{tabular}{l|ccc}
\toprule
\text{Model} & \text{AlpacaEval2: LC Win Rate (\%)} & \text{MT-Bench (1-10)} & \text{Average of Anthropic-HHH (\%)} \\ 
\midrule
\text{SFT} & 7.58 & 6.34 & xx.xx \\
\text{SPA} & 12.17 & 6.78 & 34.58 \\ 
\text{\name{} (Ours)} & 12.51 & 6.68 & \textbf{75.97} \\  
\bottomrule
\end{tabular}
}
\label{tab:diverse_dataset}
\end{table}

\noindent\textbf{Diverse backbone for Classifier.}  Table \ref{tab:Difference_backbone} shows the feature classifier trained on different backbones. We utilize a backbone from Zephyr-7b-sft-full, which is based on Mistral-7B-v0.1, instead of DeBERTa-large (304M). Interestingly, it is observed that the classifier using a larger LLM does not improve performance and even underperforms the original one using a smaller parameter. We conjecture that this inefficiency of the larger backbone is due to the small amount of training data and the long-tailed nature of preference features.
\begin{table}[htbp]
\centering
\small
\renewcommand{\arraystretch}{1.2}
\caption{\textbf{Feature classifier accuracy} with difference classifier backbone. Test accuracy of the trained feature classifier on the separately constructed test dataset.}
\resizebox{\linewidth}{!}{
\begin{tabular}{l|ccccc}
\toprule
\text{Backbone} & \text{Background} & \text{Harmlessness} & \text{Informativeness} & \text{Style} & \text{Tone} \\ 
\midrule
DeBERTa-large (original) & 0.535 & 0.512 & 0.688 & 0.496 & 0.507  \\
Mistral-7B-inst-v0.1 & 0.517 & 0.367 & 0.621 & 0.413 & 0.415 \\  
\bottomrule
\end{tabular}
}
\label{tab:Difference_backbone}
\end{table}

\noindent\textbf{Effect of Hyperparameter. } To measure the robustness of \name{}, we conduct ablation studies with different hyperparameter settings: (a) one experiment increases the number of iterations (100 $\to$ 10,000) for distribution-preserving optimization, which more aggressively modifies the distribution to follow the initial human preference distribution with a loss of information for each query; (b) the other experiment flips the system prompt sampling schedule, i.e., the first iteration is generated with temperature 0.5, increasing by 0.25 with each subsequent iteration.
As shown in Table \ref{tab:relabeling_hyper}, the proposed framework is consistently effective at mitigating preference feature bias even with significantly different hyperparameters. Additionally, the current choice of hyperparameters yields better results across all evaluation setups, which confirms its validity.

\begin{table}[htbp]
\centering
\small
\renewcommand{\arraystretch}{1.2}
\caption{Valuation results on various benchmark when hyperparameter change. (a) experiment with 10,000 iterations for distribution-preserving optimization (b) experiment with flips the system prompt sampling schedule, i.e. (0.5 $\to$ 1.25)}
\resizebox{\linewidth}{!}{
\begin{tabular}{l|ccc}
\toprule
\text{Model} & \text{AlpacaEval2: LC Win Rate (\%)} & \text{MT-Bench (1-10)} & \text{Average of Anthropic-HHH (\%)} \\ 
\midrule
SPA & 14.23 & 6.56 & 43.67 \\ 
\midrule
\text{(a)} & 12.25 & 6.81 & 68.69 \\
\text{(b)} & 11.64 & 6.63 & 62.22 \\ 
\midrule
\text{\name{} (Ours)} & \textbf{15.24} & \textbf{6.88} & \textbf{73.67} \\  
\bottomrule
\end{tabular}
}
\label{tab:relabeling_hyper}
\end{table}

\noindent\textbf{Effect of feature relabeling. } Table \ref{tab:distribution_change_relabeling} demonstrates the effect of feature relabeling, which results for sub-features chosen for each domain based on the initial percentage ratio and the largest percentage gap between sub-features generated from the classifier. It is observed that the distance from the initial distribution of human preference is largely reduced by using the proposed relabeling method.

\begin{table}[htbp]
\centering
\small
\renewcommand{\arraystretch}{1.2}
\caption{Percentage ratio of Sub-features distribution which chosen for each domain based on the largest percentage gap between sub-features generated from the classifier. This shows that feature relabeling helps to alleviate the bias caused by the inaccuracy of the classifier.}
\resizebox{\linewidth}{!}{
\begin{tabular}{l|l|rrr}
\toprule
\textbf{Domain} & \textbf{Feature} & \textbf{Original} & \textbf{Classifier} & \textbf{Relabeling} \\
\midrule
Style & Format & 36.23 & 44.45 & 34.66 \\
Tone & Formal & 38.75 & 43.18 & 36.77 \\
Harmlessness & Accuracy & 37.63 & 44.54 & 36.19 \\
Background knowledge & Novice & 10.72 & 7.23 & 12.08 \\
Informativeness & Depth & 65.47 & 82.04 & 61.39 \\
\bottomrule
\end{tabular}
}
\label{tab:distribution_change_relabeling}
\end{table}


\section{Prompts for Experiments}\label{Prompt_Set}

For the experiments, we construct prompts by modifying the ones used in \citet{lee2024aligning}:

\noindent\textbf{Feature extraction from human preference data.} Fig. \ref{fig:example_1} shows the prompts used for extracting preference features from human feedback data. 
For each class, the prompt is customized to extract a single sub-feature. 
Only for extracting preference features about the user's background Knowledge, we utilize a differently customized prompt (\ref{fig:example_3}). 

\noindent\textbf{Feature extraction from LLM's responses.} 
Fig. \ref{fig:example_3} displays the prompts used to measure the preference feature distribution of the responses from LLM. For each class, the prompt is customized to extract a single sub-feature. 
Similar to the human cases, we utilize a differently customized prompt (\ref{fig:example_4}) for the user’s background knowledge class. 

\noindent\textbf{System prompt generation.} 
Fig. \ref{fig:prompt_system_prompt} shows the prompt used to generate the system prompt based on the input preference feature set. 
This prompt takes sub-features corresponding to the five classes as input to generate the system prompt.

\begin{figure}[t!]
    \centering
    \small
    \begin{tabular}{p{0.85\linewidth}}
    \toprule 
        
\textbf{prompt} \\
\midrule 

Read the following two responses to the same prompt. After reading, determine why the preferred response is chosen over the dispreferred response, focusing on the aspect of \{domain\}. \\ \\

\textbf{Prompt:} 
[\{prompt\}] \\ \\

\textbf{Preferred Response:} 
[\{chosen\}] \\ \\

\textbf{Dispreferred Response:} 
[\{rejected\}] \\ \\

\#\#\# Question \\ 

An arbitrary person labeled the responses as preferred and dispreferred. \\
Considering the aspect of \{domain\}, what \{domain\} element does this person likely prefer? \\ \\

Select one of the following options: \\ \\

\{option\} \\ \\

Finally you have to answer as following format: \\
-Answer is \\ \\

Let's think step by step

\\ \bottomrule
    \end{tabular}
    \caption{\textbf{Prompt for feature extraction.} Input prompt for the feature extraction form pairwise preference data.}
    \label{fig:example_1}
\end{figure}

\begin{figure}[t!]
    \centering
    \small
    \begin{tabular}{p{0.85\linewidth}}
    \toprule 
        
\textbf{prompt} \\
\midrule 
Read the following two responses to the same prompt. After reading, determine why the preferred response is chosen over the dispreferred response, focusing on the aspect of the user's background knowledge. \\ \\

\textbf{Prompt:} 
[\{prompt\}] \\ \\

\textbf{Preferred Response:} 
[\{chosen\}] \\ \\

\textbf{Dispreferred Response:} 
[\{rejected\}] \\ \\

\#\#\# Question \\ 

An arbitrary person labeled the responses as preferred and dispreferred. \\
What level of background knowledge does the user have that makes them prefer the preferred response over the dispreferred response? \\ \\

Select one of the following options: \\ \\

\{option\} \\ \\

Finally you have to answer as following format: \\
- Answer is \\ \\

Let's think step by step.

\\ \bottomrule
    \end{tabular}
    \caption{\textbf{Prompt for feature extraction.} Input prompt for the feature extraction from pairwise preference data, focusing on user's background knowledge.}
    \label{fig:example_3}
\end{figure}

\begin{figure}[t!]
    \centering
    \small
    \begin{tabular}{p{0.85\linewidth}}
    \toprule 
        
\textbf{prompt} \\
\midrule 
Given a prompt and a response, analyze the response and determine which preference feature the response was likely based on. Focus on the aspect of \{domain\}. \\ \\

\textbf{Prompt:} 
[\{prompt\}] \\ \\

\textbf{Response:} 
[\{response\}] \\ \\

\#\#\# Question \\ 

An arbitrary person selected this response based on a preference for certain features within the domain of \{domain\}. \\ Considering the aspect of \{domain\}, what specific feature within this domain is the person likely prioritizing? \\ \\

Select one of the following options: \\ \\

\{options\} \\ \\

Finally, provide your answer in the following format: \\
- Answer is [selected option Alphabet] \\ \\

Let's think step-by-step.

\\ \bottomrule
    \end{tabular}
    \caption{\textbf{Prompt for feature extraction.} Input prompt for the feature extraction form single response of LLM.}
    \label{fig:example_2}
\end{figure}

\begin{figure}[t!]
    \centering
    \small
    \begin{tabular}{p{0.85\linewidth}}
    \toprule 
        
\textbf{prompt} \\
\midrule 
Given a prompt and a response, analyze the response and determine which preference feature the response was likely based on, considering the user's background knowledge. \\ \\

\textbf{Prompt:} 
[\{prompt\}] \\ \\

\textbf{Response:} 
[\{response\}] \\ \\

\#\#\# Question \\ 

An arbitrary person selected this response based on a preference for certain features related to their background knowledge. Considering the aspect of the user's background knowledge, what specific feature is the person likely prioritizing? \\ \\

Select one of the following options: \\ \\

\{options\} \\ \\

Finally, provide your answer in the following format: \\
- Answer is [selected option Alphabet] \\ \\

Let's think step-by-step.

\\ \bottomrule
    \end{tabular}
    \caption{\textbf{Prompt for feature extraction.} Input prompt for the feature extraction form single response of LLM, focusing on user's background knowledge.}
    \label{fig:example_4}
\end{figure}

\begin{figure}[t!]
    \centering
    \small
    \begin{tabular}{p{0.85\linewidth}}
    \toprule 
        
\textbf{system prompt} \\
\midrule 

You are an excellent system message generator. Read the provided rule, system message examples, and preferences carefully \\

\midrule 

\textbf{prompt} \\
\midrule 

I’m brainstorming system messages for personalizing language models. 
You are given some preferences made by human. 5 preferences are given, and each preference consists of the name of the preference and a description for it. 
Your job is to write a system message to guide a language model to behave and respond in a way that best reflects the provided human preferences. 
Please generate a creative and realistic system message. 
Refer to the given system message examples. \\ \\

\textbf{[Rule]} \\ 

- Do NOT include any greeting messages. \\
- No bullet point style. \\
- The length of the system message should not be too long. Generate a system message that is about one paragraph in length. \\
- Do not introduce any new content or task not mentioned in the preference descriptions. \\
- Do not stick to expressions like "language model", "LLM", "Assistant", and "AI" unless the preference descriptions specifically refer to language model and assistant-related content. \\
- The system message should assign a role tailored to the preferences to the model. \\ \\

\textbf{[Preferences]} \\

Style: \{style\} \\
Ensure your responses reflect the given style. \\ \\

Tone: \{tone\} \\
Make sure the tone of your responses matches the given tone. \\ \\

Harmlessness: \{harmlessness\} \\
Prioritize harmlessness as specified when crafting your responses. \\ \\

User's Background Knowledge: \{background\_knowledge\} \\
Assume the user has the background knowledge provided and tailor your responses accordingly. \\ \\

Informativeness: \{informativeness\} \\
Ensure the informativeness of your responses matches the specified level. \\ \\

\textbf{[Generated system message]} \\

\bottomrule
    \end{tabular}
    \caption{\textbf{Prompt for system prompt generation.} Input prompt for the system prompt generation from preference feature.}
    \label{fig:prompt_system_prompt}
\end{figure}

\section{Qualitative Examples} 
\noindent\textbf{System prompt sampling.} Fig.~\ref{fig:example_5} illustrates how the preference feature is sampled into the system prompt, using examples from the actual double system prompt process.
\begin{figure}[t!]
    \centering
    \small
    \begin{tabular}{p{0.85\linewidth}}
    \toprule 
        
\textbf{Preference Features} \\ 
'Format', 'Formal', 'Accuracy', 'Basic', 'Depth'
\\ \midrule 
\textbf{System prompt 1}
You are a formal and precise expert, dedicated to providing accurate and harmless information while maintaining clarity and a well-structured format. Assume the user has a basic level of background knowledge and strive to deliver detailed and in-depth responses that reflect the user's comprehension level. Your role is to ensure your responses are meticulously crafted to be both informative and non-harmful, adhering to a consistently formal tone and format
\\ \midrule 
\textbf{System prompt 2}
 You are a knowledgeable and formal advisor catering to users with basic background knowledge. Your responses should be precisely formatted, maintaining an appropriate depth of informativeness while ensuring utmost accuracy to provide harmless and useful information. Aim to deliver clear, well-structured, and respectful guidance that thoroughly addresses the queries presented
\\ \bottomrule
    \end{tabular}
    \caption{\textbf{Result of double system prompt sampling.} Two system prompts are sampled differently from the same preference features.}
    \label{fig:example_5}
\end{figure}

\noindent\textbf{Examples of generated responses.} Here, we present a direct comparison between Iterative DPO with \name{} using the generated responses on the AlpacaEval 2.0 Benchmark. 
The results are presented in Figures 
\ref{fig:response_example_3}, \ref{fig:response_example_2}, and \ref{fig:response_example_1}. 
As shown, Iterative DPO responses tend to be longer and tend to provide excessive information.

\begin{figure}[ht]
    \centering
    \small
    \begin{tabular}{p{0.85\linewidth}}
    \toprule 
        \textbf{Prompt:}   \\
What type of soil is suitable for cactus?
\\
\midrule 

\textbf{Iterative DPO response:} \\
Cacti are adapted to grow in arid and semi-arid regions with little rainfall, and therefore require well-draining soil that is low in organic matter. A suitable soil for cacti is a well-draining, porous mixture that allows water to drain away quickly and prevents waterlogging.
\\\\
Here are some characteristics of suitable soil for cacti:\\\\
1. Texture: The soil should be coarse-textured with large particles like sand and gravel, as they help with drainage. Cacti do not like their roots to be waterlogged, so a soil mixture with a lot of fine particles like clay can lead to water retention and root rot.\\\\
2. pH: Cacti prefer slightly acidic to neutral soil, with a pH range of 6.0 to 7.5. You can test your soil pH using a soil test kit or send a sample to a laboratory for analysis.\\\\
3. Nutrient Content: Cacti are not heavy feeders and do not require high nutrient content in their soil. In fact, over-fertilizing can lead to root rot or attract pests. A balanced, low-nitrogen fertilizer can be added sparingly during the growing season, but only if the plant is not getting sufficient nutrients naturally.\\\\
4. Additives: Some gardeners add horticultural sand, perlite, or vermiculite to their cactus soil to improve drainage. These additives help to loosen the soil and increase porosity, allowing excess water to drain away quickly.\\\\
5. Soil Mix: A suitable soil mix for cacti can be made by mixing one part sand or perlite with two parts coarse-textured potting soil. This mixture should be well-draining and allow excess water to escape easily.\\\\
In summary, cacti require a well-draining, porous soil mix that is low in organic matter and adequately textured to promote drainage. A balanced, low-nitrogen fertilizer can be added sparingly during the growing season, but only if the plant is not getting sufficient nutrients naturally. By providing the right soil conditions, you can help ensure healthy growth and longevity for your cacti.\\

\midrule

\textbf{\name{} response:} \\
Cactus prefer well-draining soil that is sandy, porous, and low in organic matter. A cactus soil mix should be made up of a combination of coarse sand, perlite, peat moss, and\/or vermiculite. The ideal soil pH for cactus is between 6.0 and 7.0. Avoid using heavy clay soils or organic-rich soils as they can retain too much moisture and lead to root rot. It is also important to provide good drainage by using a container with drainage holes or by placing a layer of gravel or pebbles at the bottom of the container.
\\

\bottomrule
    \end{tabular}
    \caption{\textbf{Response in AlpacaEval2.0 .} For a qualitative  comparison between Iterative DPO and \name{}, we generate the responses from the same prompt.}
    \label{fig:response_example_3}
\end{figure}

\begin{figure}[ht]
    \centering
    \small
    \begin{tabular}{p{0.85\linewidth}}
    \toprule 
        \textbf{Prompt:}   \\
Who is Larry Page?
\\
\midrule 

\textbf{Iterative DPO response:} \\
Larry Page is an American entrepreneur, computer scientist, and co-founder (along with Sergey Brin) of Google, one of the largest technology companies in the world. He was born on March 26, 1973, in East Lansing, Michigan, and grew up in the area.\\\\
Page earned his Bachelor's degree in Computer Science and Engineering in 1995 from the University of Michigan, where he later received an honorary Doctor of Science degree in 2005. After completing his studies, Page moved to Stanford University, where he pursued his PhD in Computer Science.\\\\
Page and Brin founded Google in 1998, initially as a graduate school project, with the aim of building a better search engine than those available at the time. The company quickly gained popularity and became a major player in the tech industry. Page served as Google's Chief Executive Officer (CEO) from 1999 until 2001, when he handed the reins over to Eric Schmidt.\\\\
Page has continued to play an active role in Google's development, serving as the company's CEO again from 2011 until 2019, when he stepped down in favor of Sundar Pichai. He currently serves as the CEO of Alphabet Inc., the parent company of Google, and continues to be a key figure in the company's direction and innovation.\\\\
Page is known for his focus on long-term thinking, his commitment to sustainability and renewable energy, and his dedication to improving the quality of life for people around the world through technology. He has also been recognized for his philanthropic efforts, including his work to improve education and health outcomes in developing countries through the Larry Page Center for Entrepreneurial Studies and the Page Cancer Center at Stanford University.\\

\midrule

\textbf{\name{} response:} \\
Larry Page is a co-founder of Google, a multinational technology company that provides search engine services, online advertising, and other internet-related products and services. He was born on March 26, 1973, in East Lansing, Michigan, and graduated from Stanford University with a degree in computer science and artificial intelligence. Page is known for his innovative ideas, leadership, and vision in the technology industry. He serves as the CEO of Alphabet Inc., the parent company of Google, and is one of the richest people in the world. Page is also known for his philanthropic efforts and supports various causes, including healthcare research and education.
\\

\bottomrule
    \end{tabular}
    \caption{\textbf{Response in AlpacaEval2.0 .} For a qualitative  comparison between Iterative DPO and \name{}, we generate the responses from the same prompt.}
    \label{fig:response_example_2}
\end{figure}

\begin{figure}[ht]
    \centering
    \small
    \begin{tabular}{p{0.85\linewidth}}
    \toprule 
        \textbf{Prompt:}   \\
What causes the northern lights
\\
\midrule 

\textbf{Iterative DPO response:} \\
The Northern Lights, also known as Aurora Borealis, are caused by the interaction of charged particles from the sun (solar wind) with the Earth's magnetic field and atmosphere.\\\\
When the solar wind collides with the Earth's magnetic field, it creates a region called the magnetosphere. This region traps some of the charged particles from the solar wind, including electrons and protons.\\\\
When these charged particles enter the Earth's atmosphere, they collide with the neutral atoms and molecules that make up the Earth's atmosphere, causing them to become electrically charged and emit light. The colors of the Northern Lights are determined by the type of atoms and molecules that are being ionized, as well as the altitude and energy of the collisions.\\\\
The most common colors of the Northern Lights are green, yellow, and red, but other colors such as blue, purple, and orange can also be seen under certain conditions. The brightness and movement of the Northern Lights depend on the intensity of the solar wind and the strength of the Earth's magnetic field at that particular location and time.\\

\midrule

\textbf{\name{} response:} \\
The northern lights, also known as aurora borealis, are caused by the interaction between charged particles from the sun (primarily electrons and protons) and the Earth's magnetic field and atmosphere. When these charged particles enter the Earth's magnetic field, they are deflected towards the polar regions. As they collide with the Earth's atmosphere, they excite and ionize the gas molecules, causing them to emit light in the form of colorful streaks and shapes. The colors of the northern lights depend on the type of gas molecule that is being excited and the energy of the particles colliding with them. The most common colors are green, yellow, red, blue, and purple. The intensity and frequency of northern lights depend on the solar wind, the strength and direction of the Earth's magnetic field, and other geomagnetic factors.

\\

\bottomrule
    \end{tabular}
    \caption{\textbf{Response in AlpacaEval2.0 .} For a qualitative  comparison between Iterative DPO and \name{}, we generate the responses from the same prompt.}
    \label{fig:response_example_1}
\end{figure}

\section{Use of AI Assistants} 
In preparing this work, the use of AI assistance was limited to editorial enhancements without influencing the originality or scientific contributions of the paper, such as refining sentence structure, correcting grammatical errors, and enhancing readability. 
Therefore, it is ensured that the technical content, methodology, and experimental findings remained entirely authored by the researchers. 
Likewise, the coding AI assistant helped the authors to improve the implementation of the code used in the experiment.

\end{document}